\documentclass[runningheads]{llncs}
\usepackage{graphicx}

\usepackage{tikz}
\usepackage{comment}
\usepackage{amsmath,amssymb} %
\usepackage{color}

\usepackage[accsupp]{axessibility}  %

\usepackage{xcolor}
\usepackage{array}
\usepackage{xspace}

\def\eg{\emph{e.g}\onedot}

\def\signed #1{{\leavevmode\unskip\nobreak\hfil\penalty50\hskip2em
  \hbox{}\nobreak\hfil -- #1%
  \parfillskip=0pt \finalhyphendemerits=0 \endgraf}}
\newsavebox\mybox

\definecolor{ImproveGreen}{rgb}{0.2235, 0.7094, 0.2902}

\usepackage{color}
\definecolor{crimson}{rgb}{0.86, 0.08, 0.24}
\definecolor{gray}{rgb}{0.5,0.5,0.5}
\definecolor{green}{rgb}{0, 0.4, 0}
\definecolor{orange}{rgb}{1, 0.5, 0}
\definecolor{mahogany}{rgb}{0.75, 0.25, 0.0}
\definecolor{purple}{rgb}{0.6, 0, 0.6}
\definecolor{darkgreen}{rgb}{0, 0.4, 0}
\definecolor{frenchblue}{rgb}{0.0, 0.45, 0.73}
\definecolor{red}{rgb}{1,0,0}
\definecolor{yellow}{rgb}{1,1,0}
\definecolor{magenta}{rgb}{1,0,1}
\definecolor{pink}{rgb}{1,0.412,0.706}

\newcommand{\mycomment}[1]{}

\makeatletter
\DeclareRobustCommand\onedot{\futurelet\@let@token\@onedot}
\def\@onedot{\ifx\@let@token.\else.\null\fi\xspace}

\def\eg{\emph{e.g}\onedot}

\def\etal{\emph{et al}\onedot}
\makeatother

\def\eg{e.g.,~}               %

\newlength\paramargin
\newlength\figmargin
\newlength\subfigmargin
\newlength\secmargin
\newlength\subsecmargin
\newlength\tabmargin
\newlength\eqmargin

\newcommand{\figref}[1]{Figure~\ref{fig:#1}}

\long\def\ignorethis#1{}
\definecolor{crimson}{rgb}{0.86, 0.08, 0.24}
\definecolor{green}{rgb}{0, 0.5, 0.25}
\definecolor{purple}{rgb}{0.75, 0, 1}
\definecolor{orange}{rgb}{1, 0.5, 0.25}
\definecolor{yellow}{rgb}{1, 1, 0}
\definecolor{new_blue}{rgb}{0, 0.5, 1}

\newcommand{\dafargmin}[1]{\begin{array}{c}\mbox{argmin}\\{#1}\end{array}}

\begin{document}
\pagestyle{headings}
\mainmatter
\def\ECCVSubNumber{100}  %

\title{JoJoGAN: One Shot Face Stylization}
\author{Min Jin Chong \and
  D.A. Forsyth}

\institute{University of Illinois at Urbana-Champaign\\
\email{mchong6@illinois.edu}
\email{daf@illinois.edu}}

\titlerunning{JoJoGAN}
\authorrunning{M-J. Chong et al.}
\maketitle

\begin{figure*}
    \centering
    \includegraphics[width=1\linewidth]{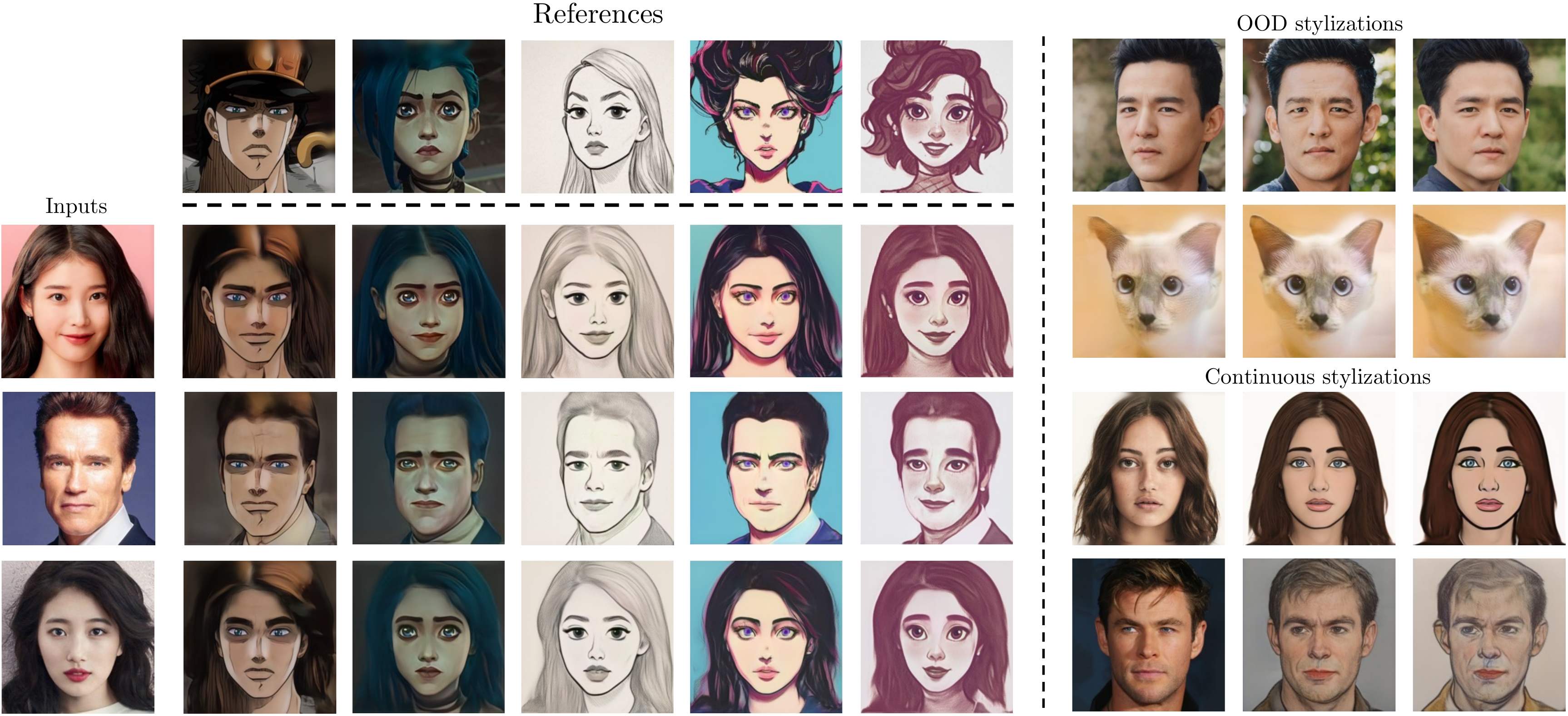}
    \caption{
      JoJoGAN accepts a single style reference image ({\bf top row}) and very quickly produces
      a style mapper that accepts an input ({\bf left column}) and applies the style
      to that input.  JoJoGAN can use extreme style references ({\bf OOD stylizations}; the cat faces are
      JoJoGAN outputs for the human inputs above.  Furthermore, JoJoGAN can apply styles to different extents ({\bf Continuous stylization});
      each row shows input; lightly stylized output; and strongly stylized output.
    }
    \label{fig:teaser}
\end{figure*}
\begin{abstract}
  A {\em style mapper} applies some fixed style to its input images (so, for example, taking faces to cartoons).  
  This paper describes a simple procedure -- JoJoGAN -- to learn a style mapper from a single example of the style.
  JoJoGAN uses a GAN inversion procedure and StyleGAN's style-mixing property to produce a substantial paired dataset from
  a single example style. The paired dataset is then used to fine-tune a StyleGAN.
  An image can then be style mapped by GAN-inversion followed by the fine-tuned StyleGAN.
 JoJoGAN needs just one reference and as little as $30$ seconds of training time.
 JoJoGAN can use extreme style references (say, animal faces) successfully. Furthermore, one can control what aspects of
 the style are used and how much of the style is applied.  Qualitative and quantitative evaluation show that
JoJoGAN  produces high quality high resolution images that vastly outperform the current state-of-the-art.
\end{abstract}

\section{Introduction}

A {\em style mapper} applies some fixed style to its input images (so, for example, taking faces to cartoons).  
This paper describes a simple procedure to learn a style mapper from a single example of the style.  Our procedure
allows, for example, an unsophisticated user to provide a style example, and then apply that style to their choice of
image.  Because stylizing face images -- make me look like JoJo -- is so desirable to unsophisticated users, we describe
our method in the context of face images; but the method applies to anything.

To be useful, a procedure for learning a style mapper should:
be easy to use; produce compelling and high quality results;
require only one style reference, but accept and benefit from more;
allow users to control how much style to transfer;
and allow more sophisticated users to control what aspects of the style get transferred.  We demonstrate with
qualitative and quantitative evidence that our method meets these goals.

Learning a style mapper is hard, because the natural method -- use paired or unpaired image translation~\cite{CycleGAN2017,huang2018munit,chong2021gans} --
isn't really practical.  Collecting a new dataset per style is clumsy, and for many styles -- Lucien Freud portraits, say --
there may not be all that many examples.  One might use few-shot learning techniques to fine-tune a
StyleGAN~\cite{Karras2019stylegan2} by adjusting the discriminator (as
in~\cite{ojha2021few-shot-gan,li2020few,robb2020fsgan,mo2020freeze}). But these methods do not have detailed supervision 
from pixel-level losses and so mostly fail to capture distinct style details and diversity.

In contrast, JoJoGAN (our procedure) takes a reference image (or images -- but one image is enough) and makes a paired
dataset using GAN inversion and StyleGAN's style-mixing property.  This paired dataset is used to fine-tune StyleGAN
using a novel direct pixel-level loss.  The mechanics are straightforward: we can obtain a mapper (and so
a rich supply of stylized portraits) from a single reference image in under a minute.
JoJoGAN can use extreme style references (say, animal faces) successfully. Natural procedures control what aspects of
the style are used and how much of the style is applied.
Qualitative examples show that
the resulting images look much better than alternative methods produce.  Quantitative evidence strongly supports our
method.   
Training and demo code is available at \url{https://github.com/mchong6/JoJoGAN}.
\section{Related Work}

{\bf Style transfer} methods likely start with~\cite{Efros01,Imageanalogies}; these are one shot methods, but do not
result in style mappers in any natural way. Neural style transfer (NST) methods start with~\cite{Gatys_2016_CVPR}; Johnson \etal offer
a learned mapper, trained with a large dataset and Gatys \etal's procedure to stylize~\cite{Johnson2016Perceptual}. In contrast, our method uses much less
data and produces much higher resolution images.
A rich literature has followed, but general style transfer methods (for example~\cite{huang2017arbitrary,WCT-NIPS-2017,park2019arbitrary}) cannot benefit from the detailed semantic and structural information captured by a GAN. 
Style transfer evaluation is mostly qualitative, but see~\cite{Yeh_2020_WACV}.
Deformable Style Transfer (DST)~\cite{Kim20DST} corrects structural errors by estimating spatial warps,
then performing traditional neural style transfer; DST achieves impressive one-shot stylization, but
warp estimation errors have significant effects and are hard to avoid (\figref{comparisons}).

{\bf StyleGAN}~\cite{karras2019style,Karras2019stylegan2}
remains the state-of-the art unconditional generative model due to its unique style-based architecture.
StyleGAN's AdaIN modulation layers (originally from~\cite{huang2017arbitrary}) have been shown to be disentangled and
exhibit impressive editability~\cite{shen2020interfacegan,shen2021closed,Chong_2021_ICCV}. StyleGAN has also been used
as a prior for numerous tasks such as superresolution~\cite{menon2020pulse} and face
restoration~\cite{wang2021towards}. Pinkney \etal~\cite{pinkney2020resolution} first showed that finetuning the
StyleGAN on a new dataset and performing layer swapping allows the StyleGAN to learn image to image translation with a
relatively small dataset. But even obtaining a small paired dataset is hard:
collection is difficult and expensive; one needs a new dataset for each new style;
and in some cases (for example, Lucien Freud portrait style) there won't be many style images in the first place.
In contrast,  JoJoGAN creates a paired dataset from a single style reference by manipulating a pretrained StyleGAN2~\cite{Karras2019stylegan2}
and a GAN inversion procedure, then finetunes using the created dataset.

{\bf One shot} learning covers many applications (detection; classification; image synthesis), and
methods remain specialized to their application. This paper focuses on one-shot image stylization, with a particular
emphasis on faces.

{\bf One shot face stylization} is now established. Learning a style mapper from very few examples results in
overfitting problems.  To control overfitting, \cite{li2020few,ojha2021few} introduce regularization terms while
\cite{robb2020fsgan,mo2020freeze} enforces constraints in the network's weights.  These methods need tens to hundreds
of style example images; in contrast, JoJoGAN works with one.  Furthermore, these methods have difficulty capturing small style
details,  likely because they rely on an adversarial loss.  BlendGAN~\cite{liu2021blendgan} introduced a
VGG-based style encoder and a weight blending module to learn arbitrary face stylization over a large styled faces
dataset. As our comparisons show, this method fails to capture small but pertinent style details in face images.
StyleGAN-NADA~\cite{gal2021stylegannada} uses CLIP~\cite{radford2021learning} to perform zero/one shot
image stylization based on text/image prompts, resulting in very strong generalization; as our comparisons show,
StyleGAN-NADA fails to capture minute facial details that are important for face stylization.

Most similar to JoJoGAN is work by Zhu \etal~\cite{zhu2022mind} (detailed
experimental comparison in \figref{comparisons_zhu_small} and Appendix);
this also uses uses GAN inversion to find a corresponding real face from a reference, so creating a paired datapoint.
Zhu \etal use this simple datapoint and a number of CLIP-based losses (from~\cite{radford2021learning}).
In contrast, JoJoGAN creates a large dataset of paired datapoints from a single one,
and so needs only a simple pixel loss (with an optional identity loss).
Zhu \etal use gradient descent inversion II2S (from~\cite{zhu2020improved}), which is slow but more accurate. In
contrast, JoJoGAN uses feed forward inversion based on a simple  encoder.
Complex losses and slow inversion procedures mean Zhu \etal require some
$15$ minutes to train on a Titan XP; in contrast, JoJoGAN require $1$.

\section{Methodology}\label{sec:methods}
\begin{figure*}[ht!]
    \centering
    \includegraphics[width=1\linewidth]{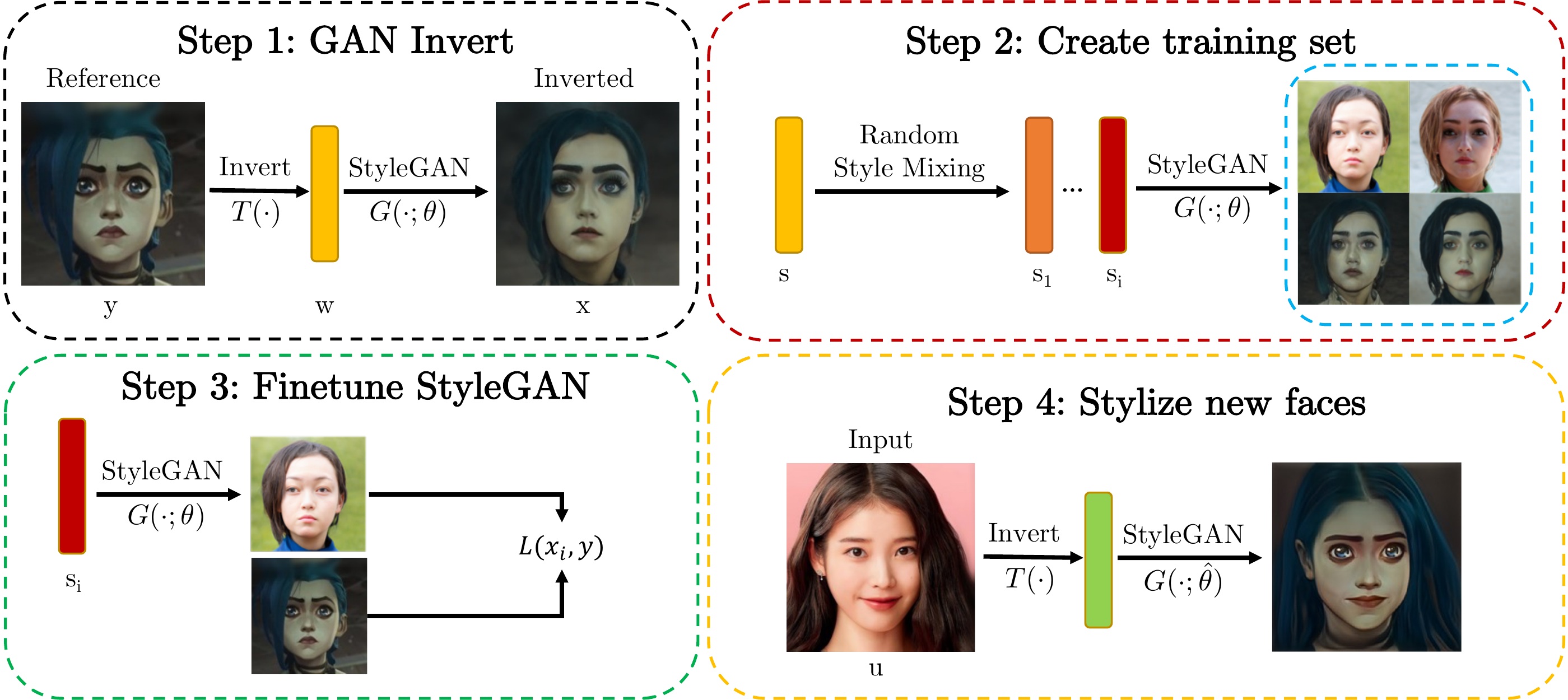}
    \caption{
    \textbf{Workflow:}
    JoJoGAN's steps are: \textbf{GAN Inversion} to obtain a code $s$ from the style reference;
    creating a \textbf{training set} $\mathcal{S}$ of similar $s_i$ via random style mixing;
    \textbf{finetuning} a StyleGAN to obtain $\hat{\theta}$ so that $G(w_i;\hat{\theta})\approx y$ using our perceptual loss;
    and \textbf{inference} by computing $G(T(u);\hat{\theta})$ for input $u$.
    }
    \label{fig:workflow}
\end{figure*}

Write $T$ for GAN inversion, $G$ for StyleGAN, $s$
for style parameters in StyleGAN's $\mathcal{S}$-space (notation after \cite{wu2021stylespace};
mixing in $\mathcal{S}$-space works better, see Appendix~\ref{section:style_mix_space}),
and $\theta$ for the parameters of the vanilla StyleGAN.
JoJoGAN uses four steps (\figref{workflow}):
\begin{enumerate}
\item {\bf GAN inversion:} We GAN invert the reference style image $y$
  to obtain a style code $w=T(y)$ and from that a set of $s$ parameters $s(w)$.
\item {\bf Training set:} We use $s$ to find a set of style codes $\mathcal{S}$ that are ``close'' to $s$. 
    Pairs $(s_i, y)$ for $s_i\in\mathcal{S}$ will be our paired training set.
  \item {\bf Finetuning:} We finetune the StyleGAN to obtain $\hat{\theta}$ such that $G(s_i;\hat{\theta}) \approx y$.
  \item {\bf Inference:} For input $u$, our stylized face is $G(s(T(u));\hat{\theta})$ (so $G\circ s \circ T$ is our style mapper).
\end{enumerate}

\paragraph{Step 1: GAN Inversion:}  
Remarkably, for any but extreme face style references $y$,  we have $G(s(T(y));\theta)$ is a realistic -- rather than
stylized -- face image (eg \figref{workflow}, step 1). This is likely because a GAN inverter is trained to produce codes that result in
realistic faces, does not see stylized faces in training, and so fails to generalize properly -- in this context, a useful property.  

\paragraph{Step 2: Training set:} We must find a set of style codes $\mathcal{S}$ that are ``close'' to $s(w)$. 
We use StyleGAN's style mixing mechanic.  We use a $1024$ resolution StyleGAN2 with $26$ style modulation
layers, so $s \in \mathbb{R}^{26 \times 512}$.  Write $M \in \{0, 1\}^{26}$ for a fixed mask,
$FC$ for the style mapping layer of the StyleGAN and $z_i \sim \mathcal{N}(0,I)$.
We produce new style codes using
\begin{equation}
    s_i = M \cdot s + (1-M) \cdot s(FC(z_i)) \label{eq:sample_w}
  \end{equation}
(and do so per batch). Different $M$ result in different stylization effects (Section~\ref{section:trainset}).

\paragraph{Step 3: Finetuning StyleGAN:}
We now assume that a properly trained style mapper will map $s_i \in \mathcal{S}$ to $y$.
This assumption certainly works, and is reasonable when the style mapper ``reduces information'' -- so, for example,
mapping faces with slightly different eye sizes or hair textures to the same reference image.
We finetune StyleGAN to obtain
\begin{equation}
    \hat{\theta}=\dafargmin{\theta} \textrm{loss}(\theta) = \dafargmin{\theta} \frac{1}{N}\sum_{i}^N \mathcal{L}(G(s_i;\theta), y) \label{eq:loss}
\end{equation}
where $\mathcal{L}$ is a novel perceptual loss (this choice is important; Section~\ref{section:loss}).

\paragraph{Step 4: Inference:}
For input $u$, our stylized face is $G(s(T(u));\hat{\theta})$ (so $G\circ s \circ T$  is our style mapper).
We could also generate random stylized samples by sampling random noise and generating with
our finetuned StyleGAN.
\begin{figure*}[t!]
    \centering
    \includegraphics[width=\linewidth]{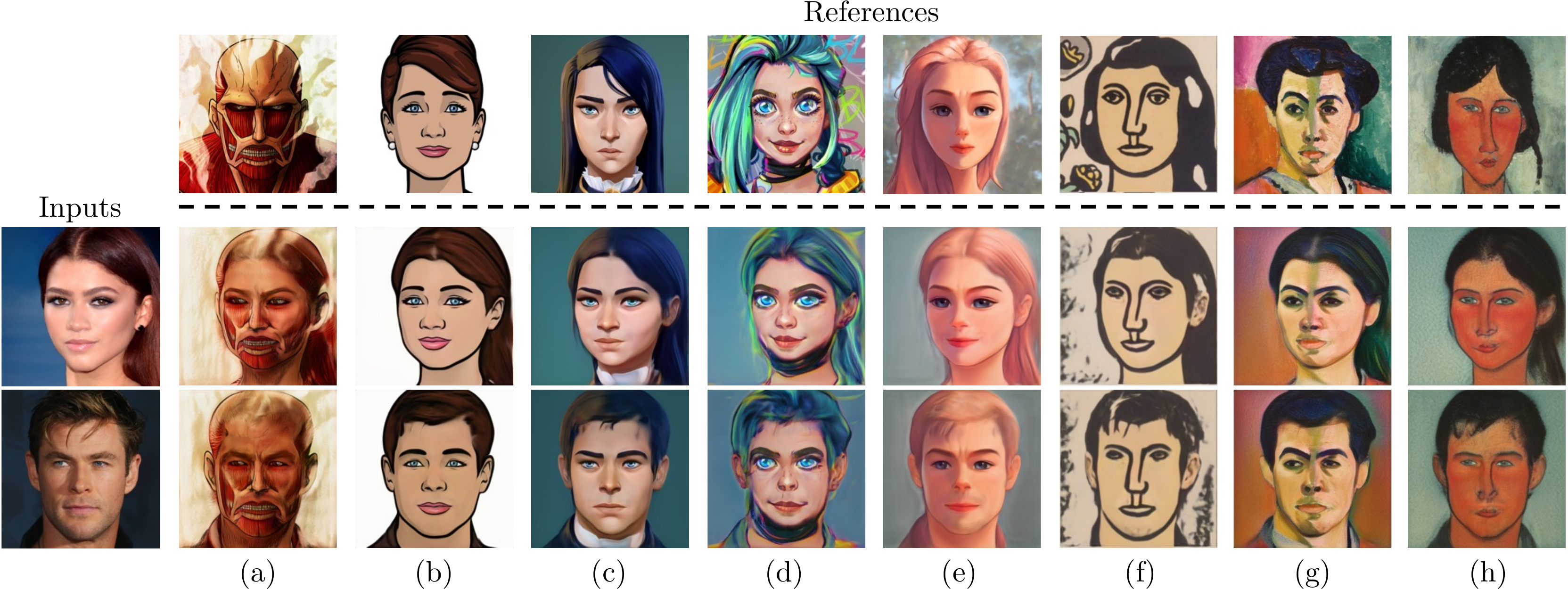}
    \caption{JoJoGAN takes a single style reference image and produces a style mapper
      (reference images on top row; inputs far left).  Note: clear following of input gender;
      subtle style details transferred (chin dimples in c; lip specularities in d);
      style lighting preserved (c, e);
      strong style effects in output, even from difficult styles (f, g, h); style
      idiosyncracies preserved (muscle fiber in a; bent nose in h; earrings in b).
    }
    \label{fig:teaser2}
    \vspace{\figmargin}
\end{figure*}

\subsection{Perceptual loss}\label{section:loss}
\begin{figure}[t!]
    \centerline{
    \includegraphics[width=0.5\linewidth]{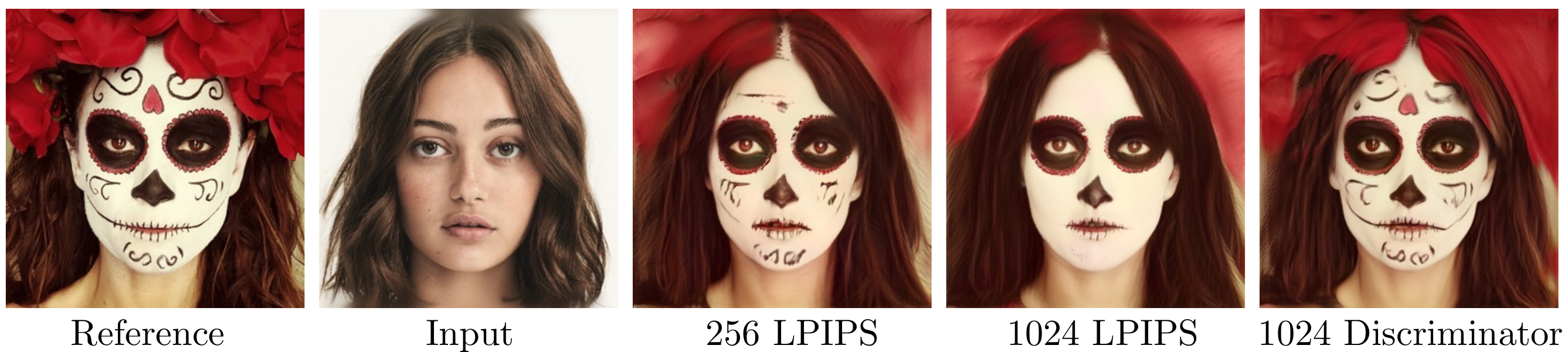} \includegraphics[width=0.5\linewidth]{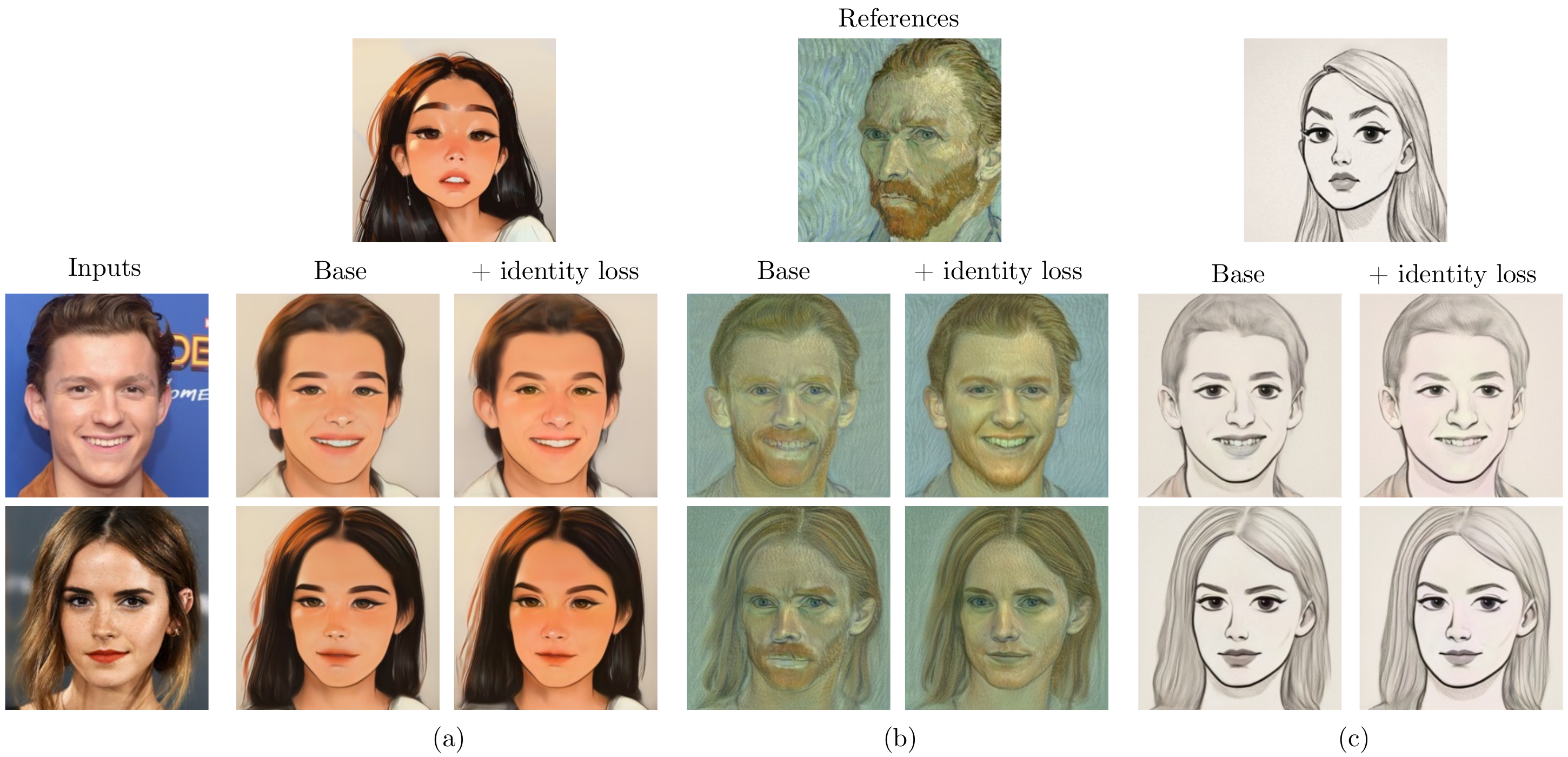}}
    \caption{{\bf Left:} The choice of loss is important (this example is typical). For the style reference and the face shown, we
      train JoJoGAN using different losses.  LPIPS at resolution $256$ resolution leads to a loss of detail
      due to downsampling. LPIPS at $1024$ does not control details, as the VGG filters (trained at $224$) are not adapted to this scale.
      We match activations at layers of the pretrained discriminator from FFHQ-trained
      StyleGAN to compute a perceptual loss that preserves detail better.
      {\bf Right:} some style inputs can result in outputs that lose identity (beard in b, for example).
      A straightforward identity loss can successfully control this effect, details in text.
    }
    \label{fig:losses}
    \label{fig:identity}
    \vspace{\figmargin}
\end{figure}

The choice of loss in Equation~\ref{eq:loss} is important (\figref{losses}). While LPIPS~\cite{zhang2018perceptual} is a natural
choice, it produces methods that lose detail.  LPIPS is built on a VGG~\cite{Simonyan15} backbone trained at a $224 \times 224$ resolution,
but StyleGAN produces $1024 \times 1024$ images.  The standard way to handle this mismatch
is to downsample the images to $256 \times 256$ before computing LPIPS~\cite{Karras2019stylegan2,tov2021designing,alaluf2021restyle}.
But this downsampling means we cannot control fine-grained details, which are mostly lost. Similarly,
computing LPIPS at the native $1024$ resolution leads to a complete loss of fine-grained detail as the VGG filters are not adapted to this resolution.

The pretrained StyleGAN discriminator is trained at the same resolution as the generator. The training process
means that discriminator computes features that do not ignore details (otherwise the generator could produce low detail images). Discriminator
features are known to stabilize GAN training when averaged over batches~\cite{salimans2016improved}.  We choose to use the difference in
discriminator activations at particular layers, per image (details in Appendix~\ref{section:feat_loss}). Write $D(\cdot)$ for the activations; then
$\mathcal{L}(G(s_i; \theta), y)= \mid\!\mid\!D(G(s_i;\theta))-D(y)\mid\!\mid_1$.
A version of this loss is used in GPEN~\cite{Yang2021GPEN} but
to our knowledge, we are the first to compare it with others and show how effective it is.

\section{Variants}
\label{section:trainset}
\label{sec:extreme}

{\bf Controlling Identity:}
Some style references distort the original identity of the inputs (\figref{identity}).
In such cases,  writing $sim$ for cosine similarity and $F$ for a pretrained face embedding network (we use ArcFace~\cite{deng2019arcface}), we use
\begin{equation}
    \mathcal{L}_{id} = 1 - sim(F(G(s_i; \theta)), F(G(s_i; \hat{\theta})))
\end{equation}
to compel the finetuned network to preserve identity; we use it only
for references that severely distort the identity and note in captions when we use it (eg. \figref{identity}(b)).

\begin{figure}[t!]
    \centering
    \includegraphics[width=1\linewidth]{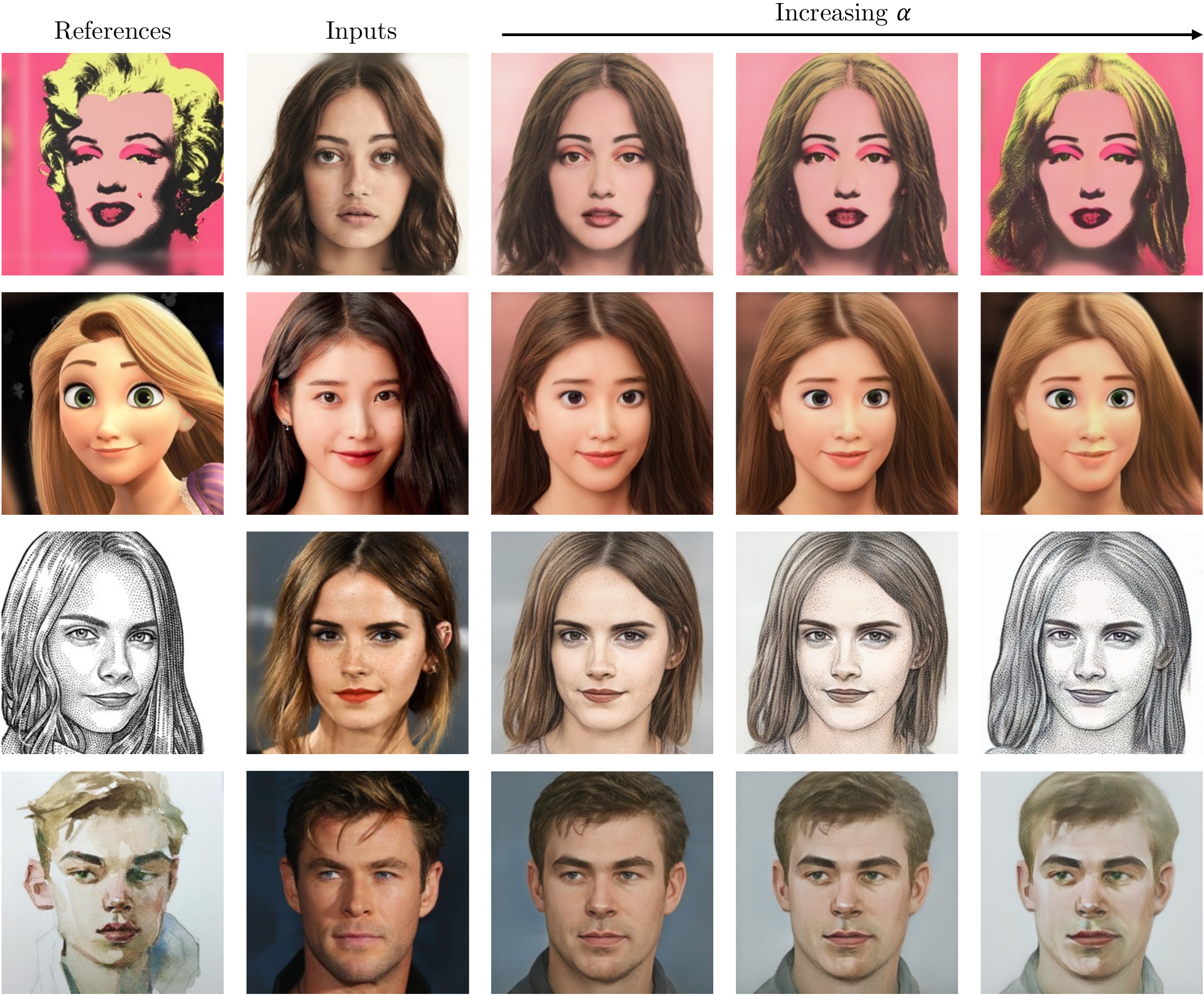}
    \caption{Feature interpolation allows a user to control style intensity.
      As $\alpha$ increases, the results take the style of the reference more strongly.
    }
    \label{fig:continous}
    \vspace{\figmargin}
\end{figure}

{\bf Controlling Style Intensity by Feature Interpolation:} Feature interpolation~\cite{chong2021stylegan} allows us to vary the intensity of the style.
Let $f_i^A$ be the layer $i$ intermediate feature maps from the original StyleGAN and $f_i^B$ from JoJoGAN; 
then we can perform continuous face stylization by using $f = (1-\alpha)f_i^A + \alpha f_i^B$ where $\alpha$ is the interpolation factor.
Increasing $\alpha$ results in stronger style intensity (\figref{continous}).

\begin{figure}[ht!]
    \centering
    \includegraphics[width=1\linewidth]{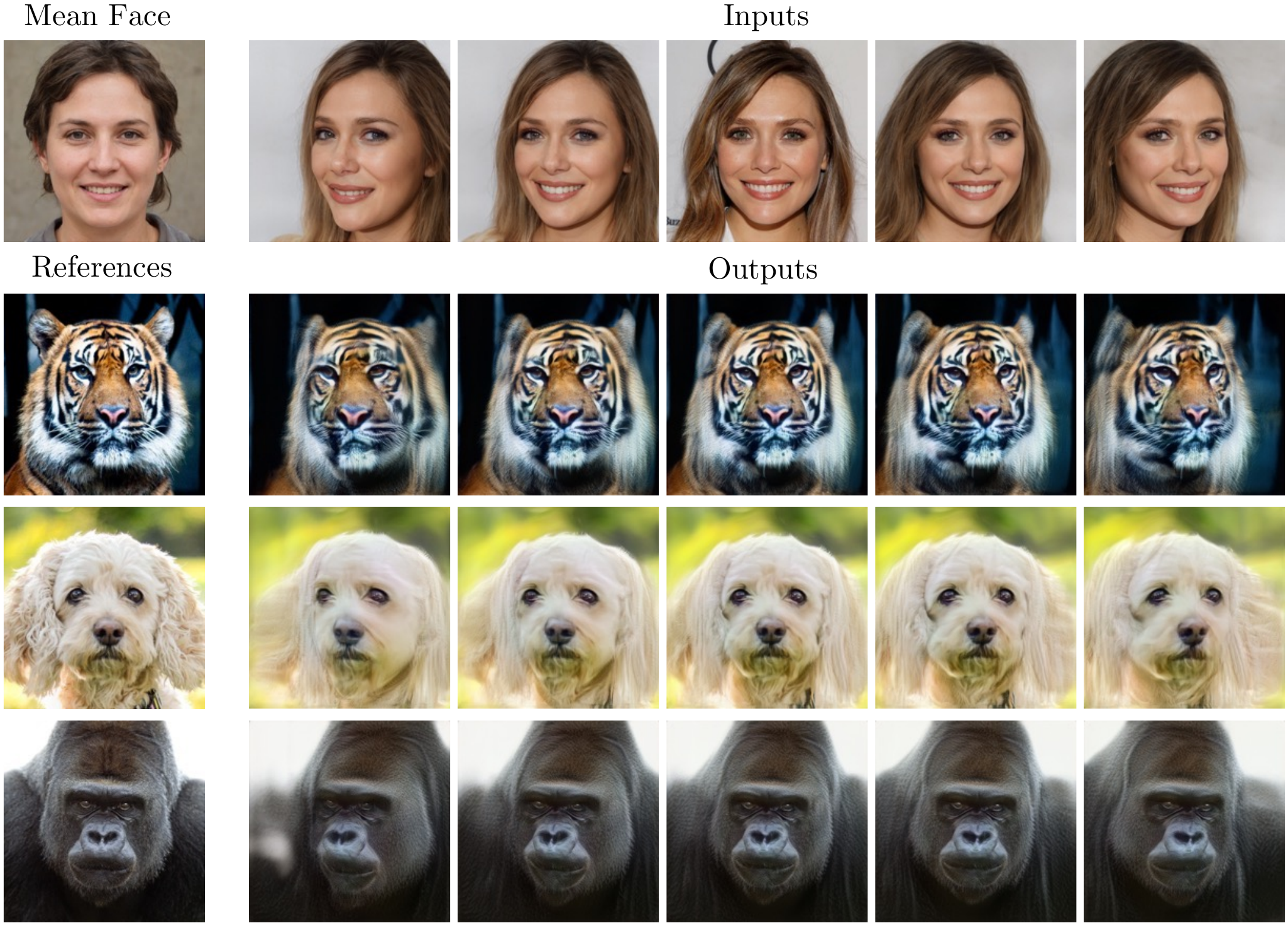}
    \caption{
    \textbf{OOD references and using $\overline{w}$:}
    JoJoGAN is able to handle OOD references that do not invert well by using mean style code $\overline{w}$.
    Even on animal faces which are semantically very different the human faces StyleGAN was trained on, JoJoGAN can
    generate realistic animal faces with poses that matches the input.
    }
    \label{fig:ood}
    \vspace{\figmargin}
\end{figure}

{\bf Extreme Style References:} For JoJoGAN to work, $\mathcal{S}$ has to consist of $s_i$ that produce sensible responses from the StyleGAN. If the
style reference is (roughly) a human face, there are no problems. An extreme style reference image is one where GAN inversion
produces $s$ that is out of distribution for the StyleGAN, for example, an image of an animal face.  We are not aware of
any test (other than trying) to distinguish between extreme and standard style references, but \figref{oodw} in the Appendix demonstrates
that using $s$ from GAN inversion on animal faces results in poor style transfer.
For extreme style references $y$, rather than use $s(T(y))$ to construct $\mathcal{S}$, we use
the mean style code $\overline{s} = \sum_1^{10000} s(FC(z \sim \mathcal{N}(0,I)))$  (note
this style code is the best possible estimate of $s(T(y))$ for an image $y$ {\em that one does not have}).
With this modification, JoJoGAN works well on extreme style references (\figref{ood}; note how the
animal head poses are controlled by the input images).

\begin{figure}[ht]
    \centering
    \includegraphics[width=1\linewidth]{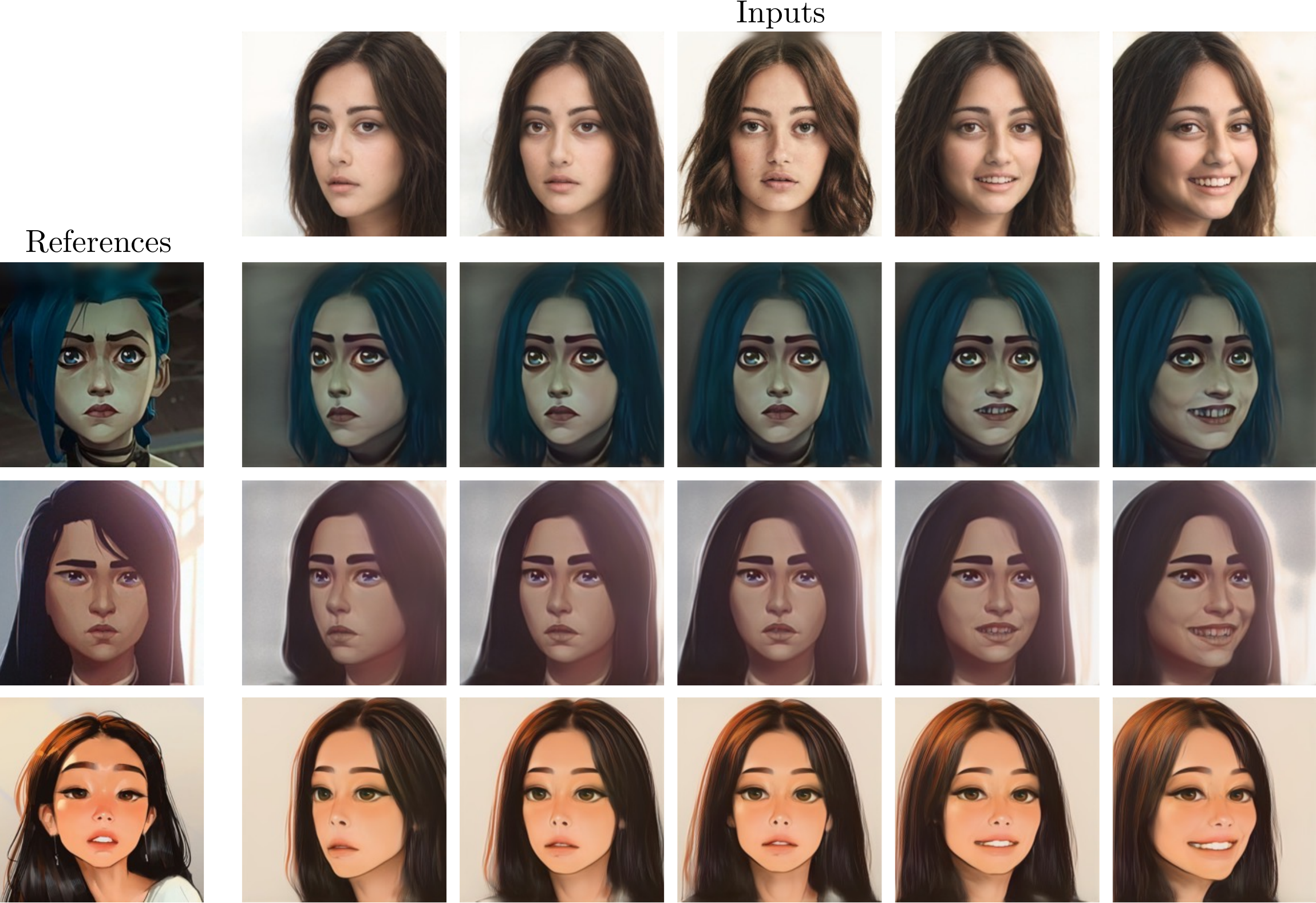}
    \caption{
    JoJoGAN produces smooth and consistent stylization as the face moves and changes expressions.
    }
    \label{fig:turn}
    \vspace{\figmargin}
\end{figure}
{\bf Multi-shot Stylization:} JoJoGAN extends to multi-shot stylization in the natural way (use each reference to construct a $\mathcal{S}_k$ for each reference $y_k$; now finetune using \[ \frac{1}{M*N}\sum_{j}^M \sum_{i}^N \mathcal{L}(G(s_{ij};\theta), y_j).\]  Using more than one
reference produces small but useful qualitative improvements in the style mapper (\figref{multi})

\section{Controlling Aspects of Style} \begin{figure}[t!]
    \centering
    \includegraphics[width=1\linewidth]{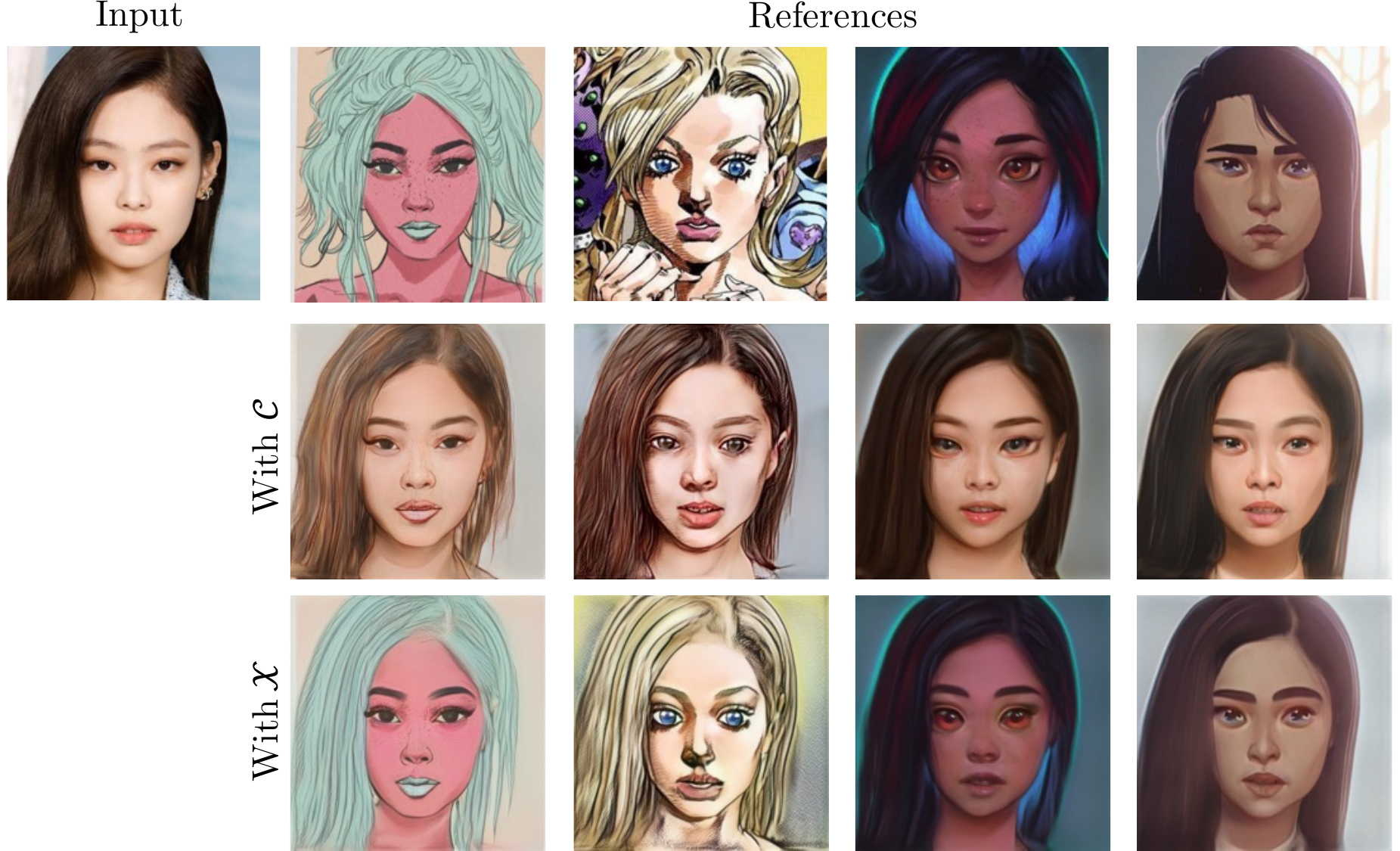}
    \caption{The aspects of style that are transferred can be chosen using $M$. The text describes our procedures
      to create two different datasets $\mathcal{C}$ and $\mathcal{X}$ from different choices of $M$ that yield different stylizations.
      Finetuning using $\mathcal{C}$ mostly preserves the colors of the input; finetuning using $\mathcal{X}$ mostly reproduces the colors of
      the style example.
    }
    \label{fig:preserve_color}
    \vspace{\figmargin}
\end{figure}

Style transfer is intrinsically ambiguous.  The output should be ``like'' the reference as to style, and ``like''
the input as to content, but the distinction between content and style is vague.  JoJoGAN offers methods to choose
whether (say) the output should have exaggerated eyes (like the reference) or more natural eyes (like the input).  Simple
control is obtained by choice of mask and by loss.  More detailed control follows by careful attention to the GAN inversion.

{\bf Controlling Aspects of Style by Mask Choice and by Loss:} Different choices of $M$ will produce significant differences in
$\mathcal{S}$, and so in results. Replacing too many elements of $s$ with random numbers may result in a JoJoGAN that maps
every face to the style reference; replacing too few means finetuning sees too few examples. Furthermore, replacing
elements at locations corresponding to different StyleGAN layers controls different effects
(see~\cite{Karras2019stylegan2}). \figref{preserve_color} demonstrates this choice has significant effects by displaying
results from two different $M$. The first gives dataset $\mathcal{X}$, the second $\mathcal{C}$.
Both masks are chosen to maintain the input face pose and hairstyles while allowing features such as eye sizes and textures
to vary, so the mask has ones in locations known to correspond to pose and zeros in those known to correspond to eye-sizes,
see~\cite{Chong_2021_ICCV}.  But $\mathcal{C}$ is chosen so that the color of the input is preserved (so ones in relevant
locations); and $\mathcal{X}$ so that color is driven by the style example. To ensure that the color of the input is
preserved for the $\mathcal{C}$ case, we apply the loss in Equation~\eqref{eq:loss} to grayscale versions of the relevant
images.  This means the StyleGAN is finetuned to obtain the spatial appearance of the style target, but not its colors
(variants in  Appendix~\ref{section:vary_dataset}) 
 
\subsection{Control by Manipulating GAN Inversion}
\begin{figure}[ht]
    \centering
    \includegraphics[width=1\linewidth]{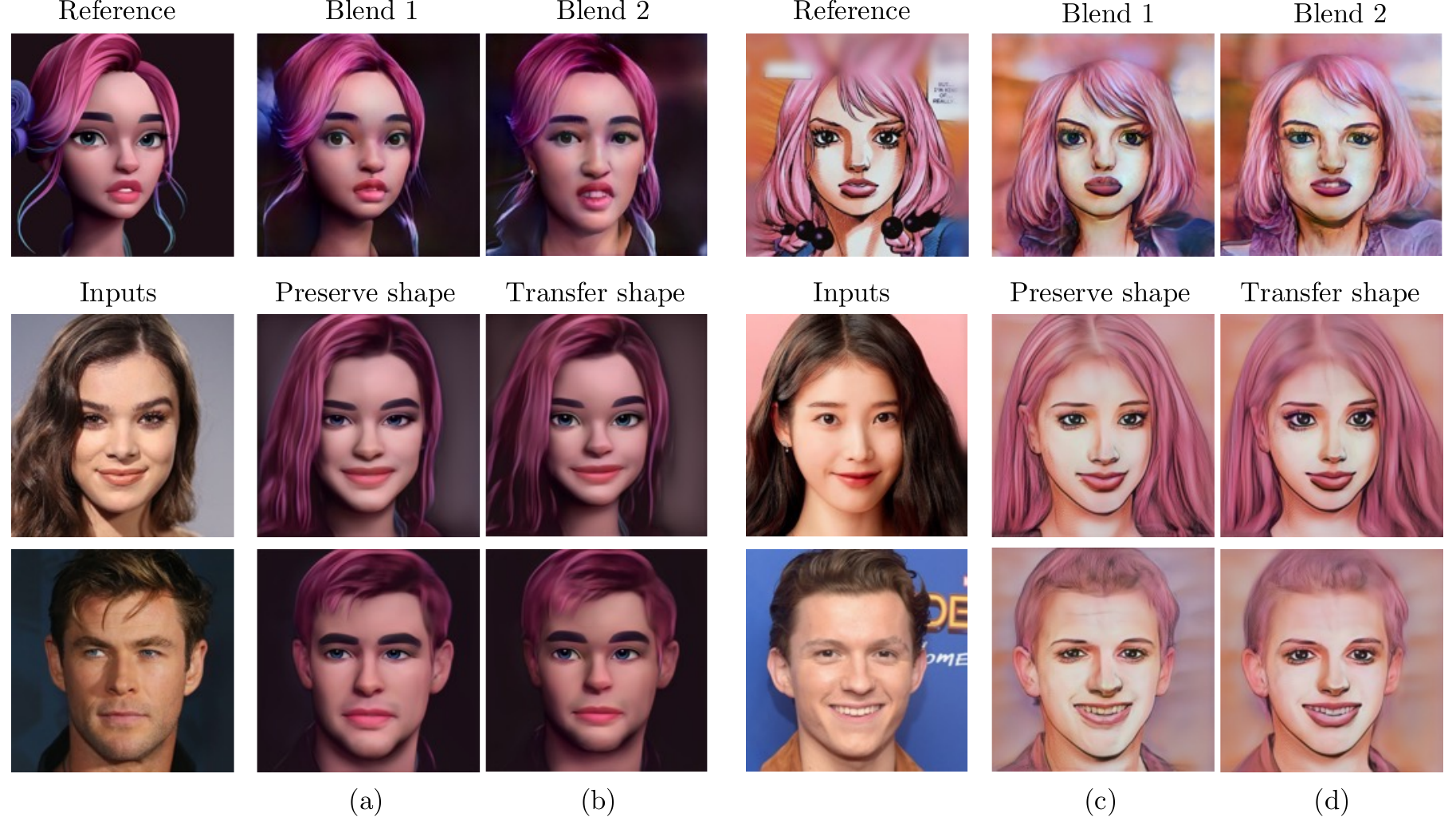}
    \caption{
      How style is transferred can be controlled by blending the codes from two GAN inverters, then applying the JoJoGAN pipeline.
      (for these examples, one inverter just produces the mean code).
    For each reference, {\bf top} shows $G(s(V(y));\theta)$ for different blends.
        Notice how blending the inverter codes produces substantial changes in the inversion (eg {\bf left} reference).
      By choice of blend, we can produce style mappers that tend to {\bf preserve} the shape of input eye, nose and face or to {\bf transfer}
      shapes from the reference.
      So, for example, (a) and (c) have eyes more like the input; but (b) and (d) have larger eyes, more like the reference.
        (b) has significantly smaller faces than (a).
    }
    \label{fig:shapes}
    \vspace{\figmargin}
\end{figure}
The choice of GAN inverter matters.  If the GAN inverter produces an extremely realistic face from the reference, JoJoGAN will be trained to map
$s_i$ that represent highly realistic faces to the style reference, and so will tend to produce aggressively stylized
faces. By the same argument, if the GAN inverter produces a somewhat stylized face from the reference, JoJoGAN will tend to produce lightly
stylized faces and to preserve the features of the input face (so an input with small eyes will result in an output with
small eyes, say -- example in Appendix \figref{inversion_comparisons}). This effect can be used to control how much and what style is transferred by blending inverted codes.

Using two GAN inverters is clumsy in practice, but recall the mean style code is the best possible estimate of $s(T(y))$ for
an image $y$ {\em that one does not have}), and so is the output of a (rather bad, but very fast) GAN inverter.  We produce a
virtual inverter $V(y)$ by blending the code produced by our standard inverter with the mean, using the procedure of Section~\ref{sec:methods}
(but a different mask $M$).  The blend is adjusted so that $G(s(V(y)); \theta)$ has desirable properties (so, for example, to preserve the eyes of the reference, $G(s(V(y)); \theta)$ should have realistic eyes).  We then apply the JoJoGAN pipeline using $V$ rather than $T$ to generate
training data.  Using $V$ rather than $T$ in training changes the pairs $(s_i, y)$ used in finetuning, and so the behavior of $G$.  At
inference, we compute $G(s(T(u); \hat{\theta})$ as before.  \figref{shapes} demonstrates the extent of our style control. In \figref{shapes}(b),
using the blended inversion gives us larger eyes and thicker lips compared to using the accurate inversion (a).
Further detail on blending the inverter in Appendix~\ref{section:gan_choice}.

\section{Experiments}

{\bf Setup:} For GAN inversion, we use ReStyle~\cite{alaluf2021restyle}.
We finetune JoJoGAN for $200$ to $500$ iterations depending on the reference with Adam optimizer~\cite{Kingma2015AdamAM}
at a learning rate of $2 \times 10^{-3}$. Finetuning on an Nvidia A40 takes about $30$ to $60$ seconds.

\begin{figure*}[ht!]
    \centering
    \includegraphics[width=0.9\linewidth]{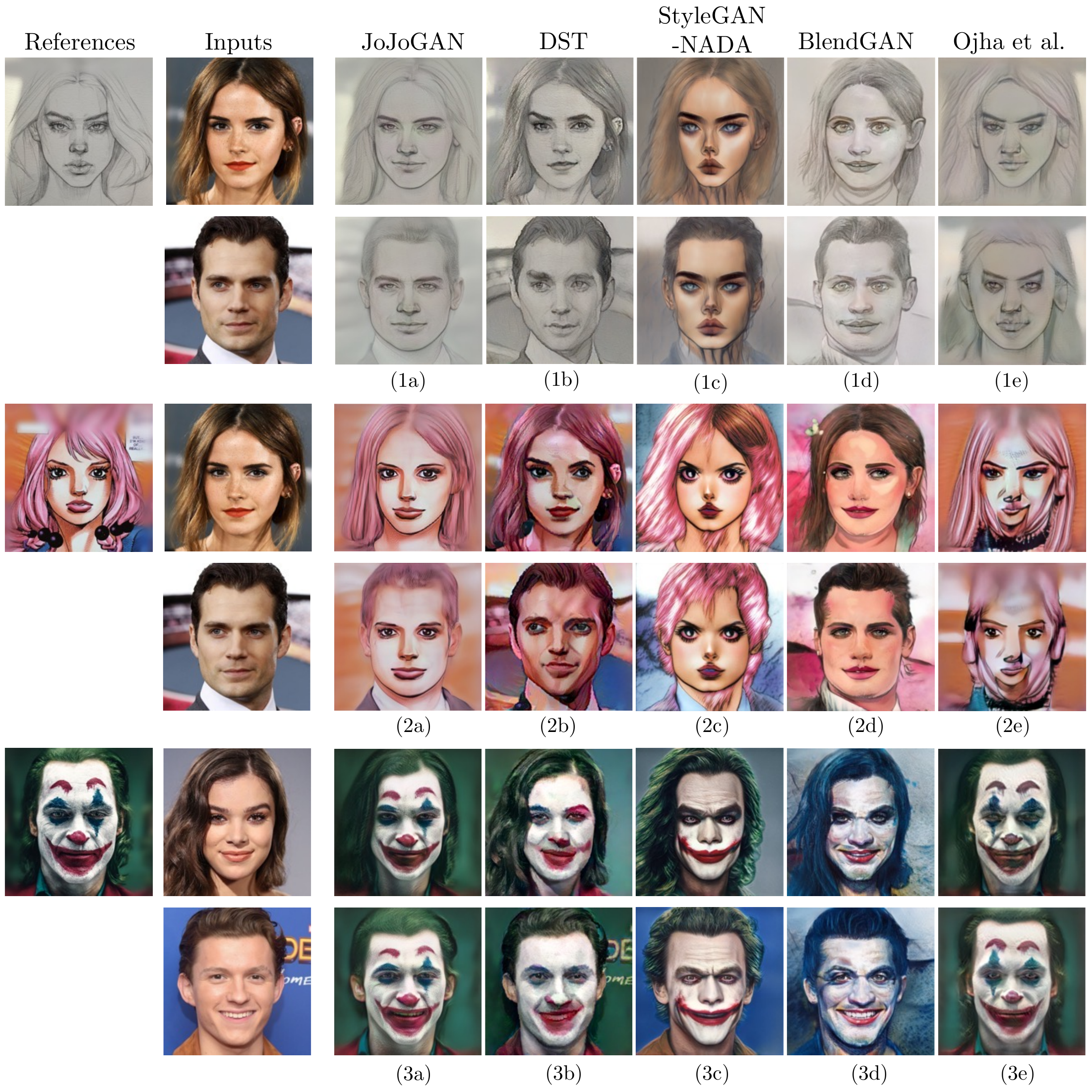}
    \caption{
      JoJoGAN offers visible qualitative improvements over current
      SOTA methods for one shot face stylization.
      JoJoGAN captures the distinctive rendering style of the reference while
      preserving input pose, expression and identity. 
      Note: excessive contrast (1b); color errors (1c, 2b, 3d); distorted facial layout (d, e);
      chin shape (b).
    }
    \label{fig:comparisons}
    \vspace{\figmargin}
\end{figure*}
{\bf Qualitative evaluation:} A style mapper should: produce good looking outputs; faithfully transfer features from the style
reference; and preserve the identity of the input.  Qualitative evaluation shows JoJoGAN has these properties and vastly outperforms current methods.

\begin{figure}[ht!]
    \includegraphics[width=1\linewidth]{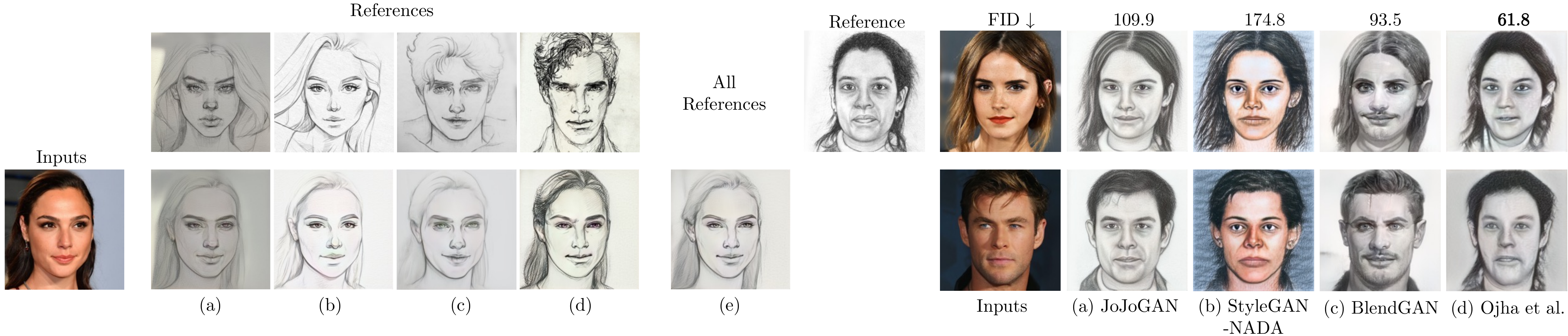}
    \caption{{\bf Left:} JoJoGAN's method extends cleanly to deal with
      multiple style references, if they are available.  The figure
      compares one-shot stylizations of a reference with
      a multi-shot stylization for one input (more in supplementary). Note aggressive copying in
      the case of a single reference, including: noses in (a);
      lips in (b) and (c); and chin dimples in (b) and (d). This effect is notably
      muted when more references are available (and
      JoJoGAN can blend details from references), so (e) mouth and chin
      follow the input more closely.
      {\bf Right:}
      JoJoGAN's FID score on the sketches dataset~\cite{ojha2021few-shot-gan}
      is significantly larger than that of the best comparison.
      BlendGAN gets a better FID, but does not capture reference
      style well (note strong shading gradients, absent from the style reference);
      Ojha \etal get the best FID, but impose strong distortions on the input
      face (note other comparisons in \figref{comparisons}).
    }
    \label{fig:fid}
    \label{fig:multi}    
    \vspace{\figmargin}
\end{figure}

{\bf Comparisons:} \figref{comparisons} shows comparisons of JoJoGAN to the
state-of-the-art one/few shot stylization methods StyleGAN-NADA~\cite{gal2021stylegannada},
BlendGAN~\cite{liu2021blendgan}, Ojha \etal~\cite{ojha2021few-shot-gan} and DST~\cite{Kim20DST}.
JoJoGAN captures small details well that define the style while maintaining the identity of the input face well.
JoJoGAN results are typically improved when there are multiple consistent style references. \figref{multi}
compares several one-shot stylizations of each of a set of examples with a multi-shot stylization using all.
Notice that one-shot stylization copies effects from the style reference aggressively (as it must), whereas when there
are multiple style examples, JoJoGAN is able to blend details to hew more closely to the input.

\begin{figure}[ht]
    \centering
    \includegraphics[width=1\textwidth]{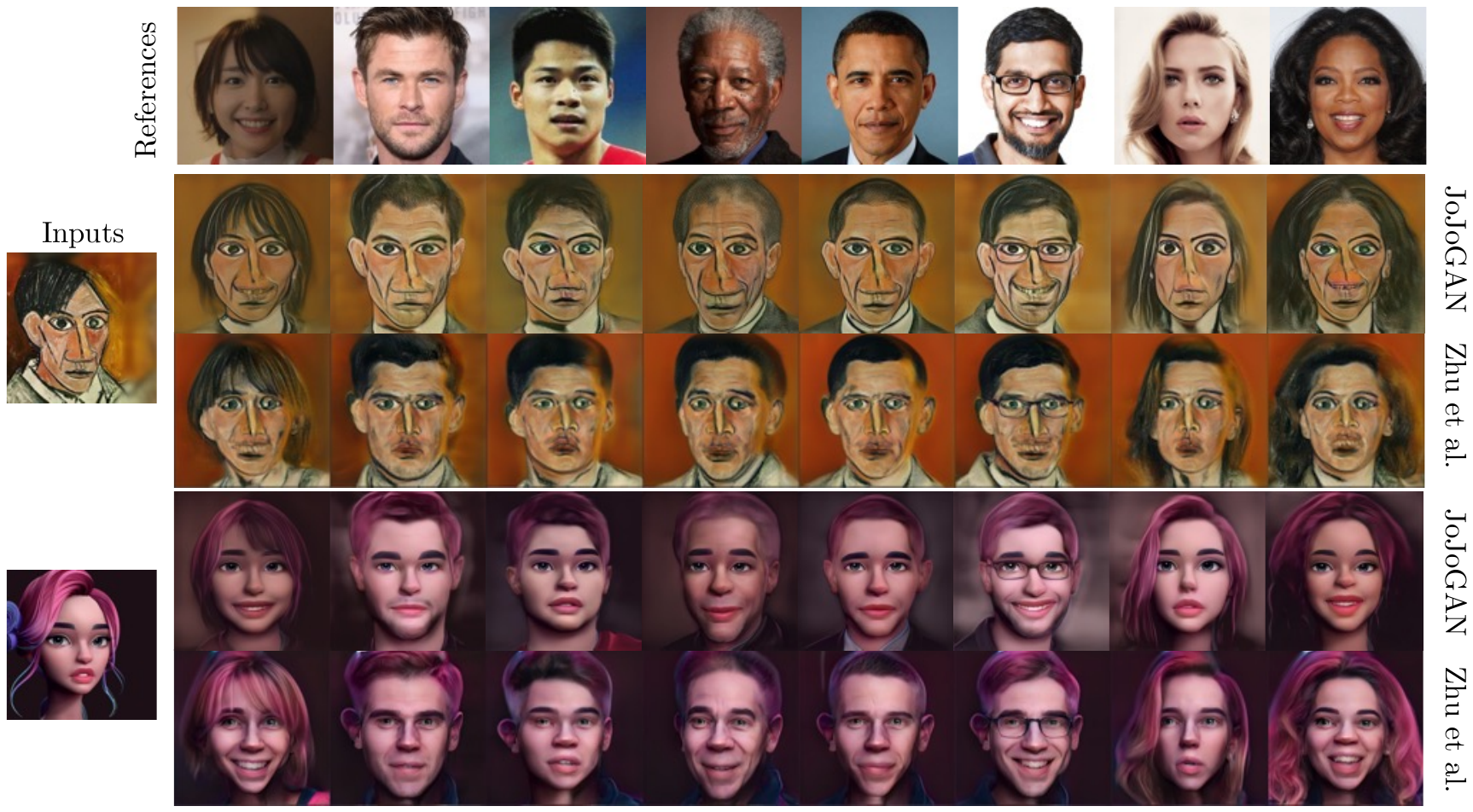}
    \caption{
      We compare with Zhu \etal~\cite{zhu2022mind} on two examples for references used in their paper and described as hard cases there    (others in supplementary). For each reference, the top row is JoJoGAN while the second row is Zhu \etal
      Note how their method distorts chin shape, while JoJoGAN produces strong outputs.
    }
    \label{fig:comparisons_zhu_small}
    \vspace{\figmargin}
\end{figure}

\figref{comparisons_zhu_small} shows a comparison with~\cite{zhu2022mind} (two examples in figure;
others -- except $2$, for which we cannot find source -- in supplementary).
Note we can use only references shown in their paper, as the method is not open sourced.

{\bf Quantitative Evaluation by User study:}
We proceed in two stages, to reduce choice fatigue for users.
From \figref{comparisons}, DST gives good results in most cases while other methods produces examples with severe problems.
We therefore compare JoJoGAN to non-DST methods in a first study, and to DST in a second.
In each, users see a style reference, an input face, and stylizations from the methods and are asked
to choose the stylization that best captures the style reference and while preserving the original identity.
The first study resulted in a total of $186$ responses from $31$ participants who overwhelmingly prefer
JoJoGAN to other methods at $80.6\%$; the effect is so large that no significance issues arise. 
The second study gathered $96$ responses from $16$ participants who prefer JoJoGAN to DST at $74\%$.

{\bf Quantitative Evaluation by FID:}  FID~\cite{heusel2017gans} is
a metric that is widely used to evaluate the quality and diversity of generated images by comparing population statistics.
FID can be used to evaluate style mappers as follows~\cite{ojha2021few}. Randomly select
a reference from the style dataset and performing one shot stylization with it;
now stylize a set of face images and compute the FID between the result and the original style dataset.
To compute FID, we perform one shot stylization using the sketches dataset~\cite{ojha2021few-shot-gan} and compute FID using the test set.
JoJoGAN scores well behind SOTA on this metric. We report FID for JoJoGAN for candor and show FID for SOTA comparisons in
\figref{fid}, but point out that FID is a poor metric for style mappers.  
The procedure described cannot measure the fidelity with which the
mapper preserves the input (for example, the FID for the
completely ineffectual mapper that just produces a random sample from the style
dataset would be close to zero).  Further, a perfect style mapper might produce a high FID with the protocol described, 
because its stylized images should be biased toward the input (for example, a perfect mapper with only male input images
should produce a population of sketches that is not close to the original set of sketches).
Finally, the datasets used for stylization are often very small (290 in the case of the sketches dataset),
and computing FID for a small dataset is dangerous due to large biases \cite{chong2020effectively}.

\begin{figure}[ht!]
    \centering
    \includegraphics[width=0.75\linewidth]{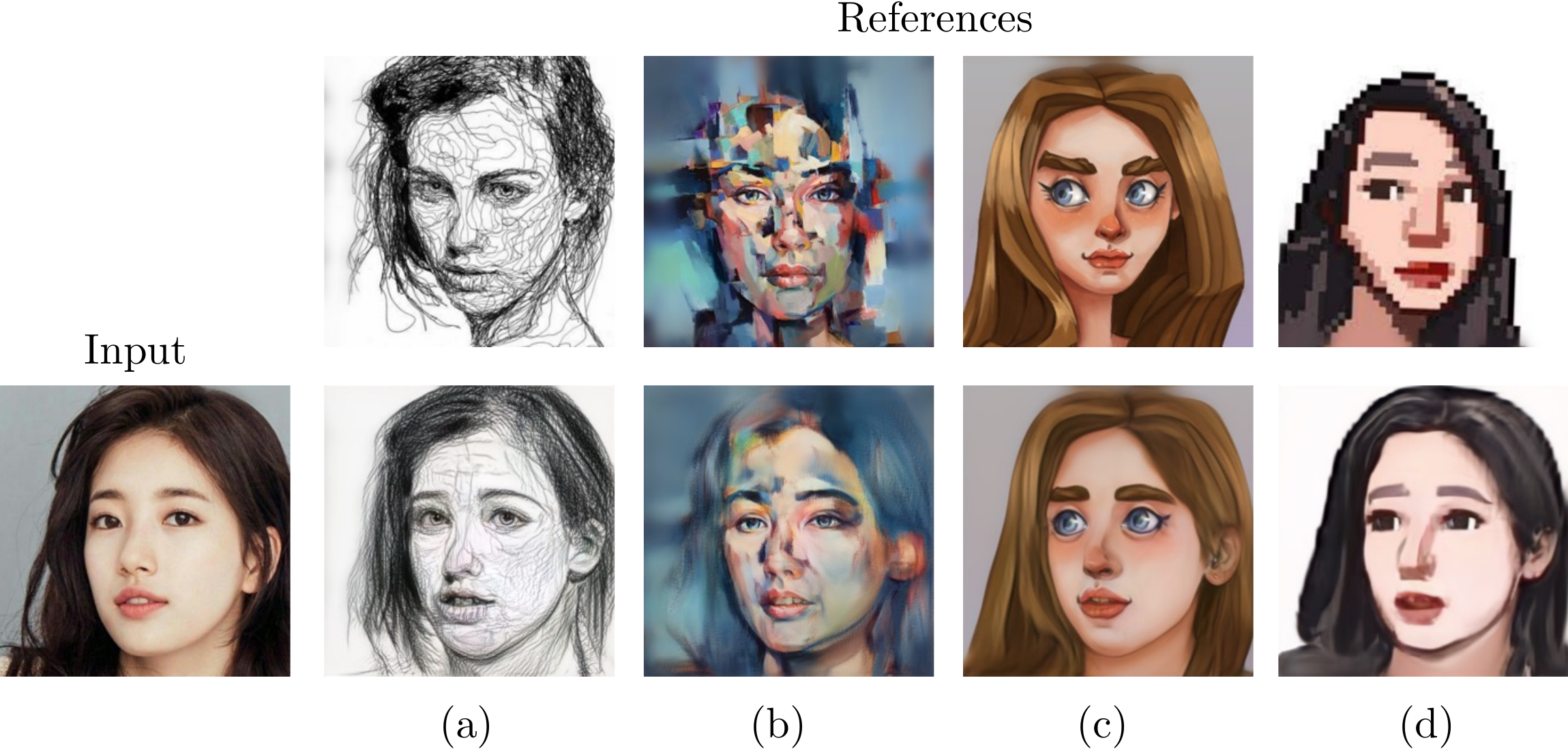}
    \caption{Some style references are hard for JoJoGAN, likely a
      result of complicated structures in the style reference that are
      unfamiliar to StyleGAN.  Note: loops in (a) mapped to strokes in the output;
      structure of brush strokes in (b) being broken up in output; gaze direction in
      (c) controlled by style reference rather than by input;
        high frequency pixel grids in (d) map to smooth strokes.
    }
    \label{fig:failure}
    \vspace{\figmargin}
\end{figure}

{\bf Failures:} Using too small a $\mathcal{S}$ leads to problems (Appendix \figref{preserve_color2}), typically
artifacts and missing style details.  As JoJoGAN only sees a single style reference, it does not always work for all style references.
One common issue JoJoGAN has is that the eye gaze direction is often driven by the reference image rather than the input.
The intended behavior is to preserve the gaze direction of the original input, yet JoJoGAN copies the reference instead.
\figref{failure} shows results on very difficult references, illustrating visual failure modes.

\bibliographystyle{splncs04}
\bibliography{bibliography}
\clearpage
\appendix
\section{Appendix}
\subsection{Choice of GAN inversion}\label{section:gan_choice}
\begin{figure}[ht]
    \centering
    \includegraphics[width=1\linewidth]{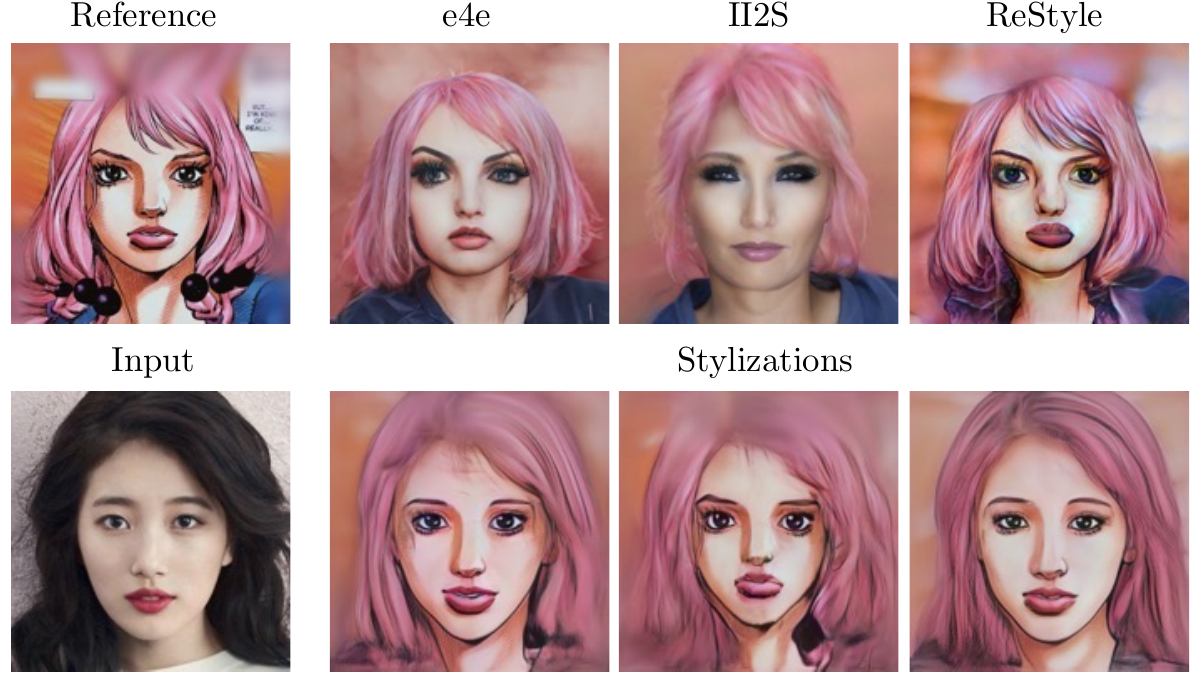}
    \caption{
    The choice of GAN inversion matters.
    We compare JoJoGAN trained on e4e~\cite{tov2021designing}, II2S~\cite{zhu2020improved}, and ReStyle~\cite{alaluf2021restyle} inversions.
    II2S gives the most realistic inversions leading to stylizations that preserves shapes and proportions of the reference.
    ReStyle gives the most accurate reconstruction leading to stylization that better preserves the features and proportions of the input.
    }
    \label{fig:inversion_comparisons}
    \vspace{\figmargin}
\end{figure}
JoJoGAN relies on GAN inversion to create a paired dataset.
We investigate the effect of using $3$ different GAN inversion methods, e4e~\cite{tov2021designing},
II2S~\cite{zhu2020improved}, and ReStyle~\cite{alaluf2021restyle} in \figref{inversion_comparisons}.

Using e4e fails to accurately recreate the style reference and conveniently gives us a corresponding real face.
On the other hand, ReStyle more accurately inverts the reference, giving a non-realistic face.
II2S is a gradient-descent based method with a regularization term that allows us to map the style code to higher density region in the latent space.
The regularization term results in a very realistic face that are somewhat inaccurate to the reference.

The different inversions give us different JoJoGAN results.
Training with ReStyle leads to clean stylization that accurately preserves the features and proportions of the input face.
Training with II2S on the other hand, leads to heavy stylization that borrows the shapes and proportions from the reference.
However, this also leads to pretty heavy semantic changes from the input face and artifacts (note the change of identity, artifacts along the neck).

\begin{figure}[ht]
    \centering
    \includegraphics[width=1\linewidth]{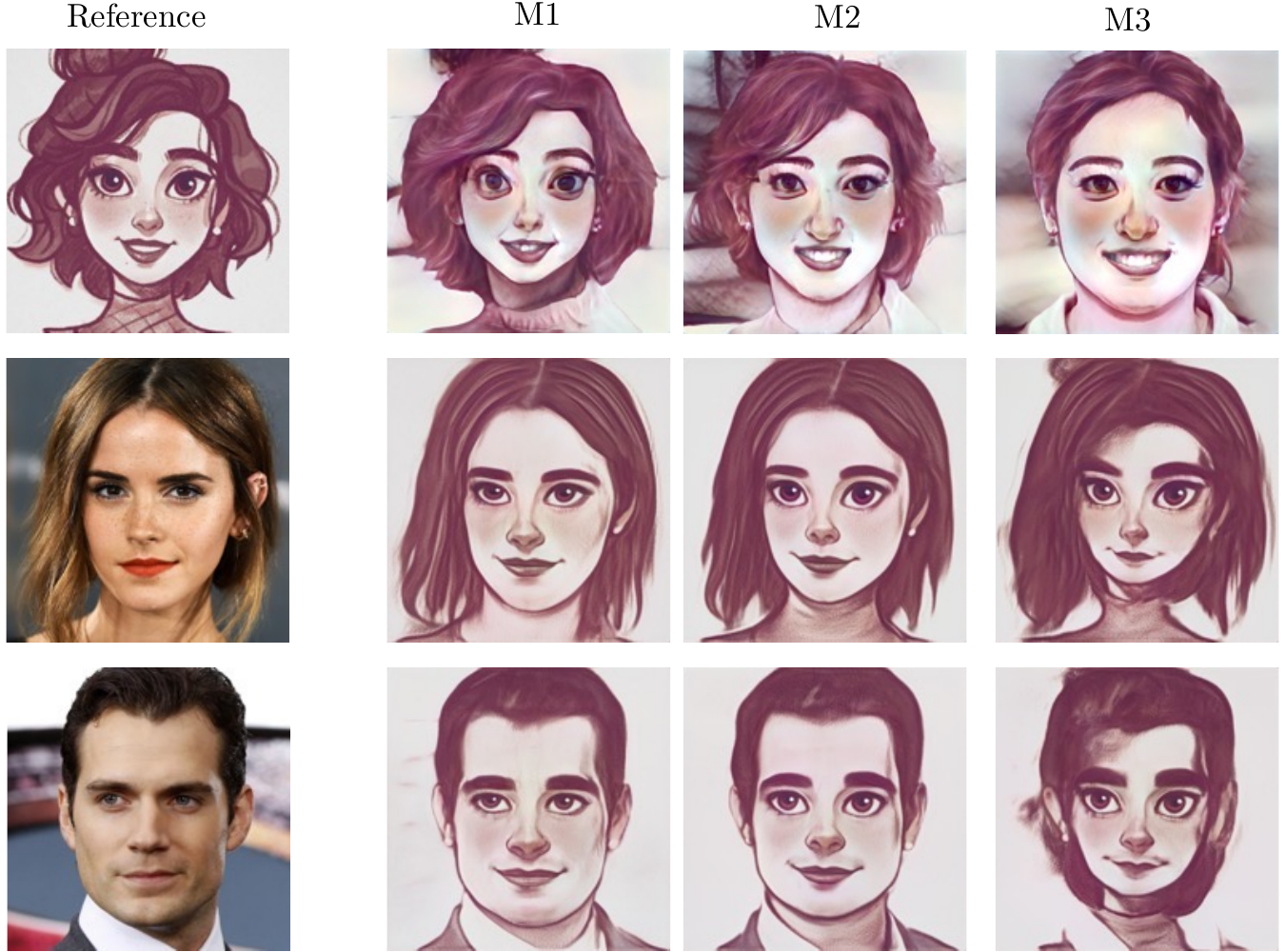}
    \caption{
    The choice of $M$ matters.
    $M$ controls the blend between the inverted style with the mean style.
    $M1$ is the closest to the reference, leading to smaller features (\eg eyes).
    $M3$ is the closest to a real face, leading to exaggerated features more like reference and also significant artifacts.
    }
    \label{fig:compare_M}
    \vspace{\figmargin}
\end{figure}
In practice, we blend the styles codes from ReStyle and the mean face.
For $M$, we borrow the style code from mean face at layers $7$, $9$ and $11$.
This borrows the facial features of the mean face to the inversion.
However, it is impossible to only affect the proportions of the features by simply blending coarsely at a layer level.
For example, naively blending the mean face can change the expression of the inversion, e.g. from neutral to smiling or introduce artifacts.
We thus have to blend at a finer scale, which we are able to do so by isolating specific facial features in the style space
using RIS~\cite{Chong_2021_ICCV}.
\figref{compare_M} compares the results of using different $M$ for blending.
Note that when the blended image is more face-like ($M3$), the exaggerated features of the reference is transferred.
However, significant artifacts are introduced, see $M3$ row $2$.
By carefully selecting $M$, we can transfer the exaggerated features while avoiding artifacts, see $M2$.

\subsection{Identity loss}
Before computing identity loss, we grayscale the input images to prevent the identity loss from affecting the colors.
The weight of the identity loss is reference dependent, but we typically choose between $2 \times 10^3$ to $5 \times 10^3$.

\subsection{Choice of style mixing space}\label{section:style_mix_space}
\begin{figure}[ht]
    \centering
    \includegraphics[width=1\linewidth]{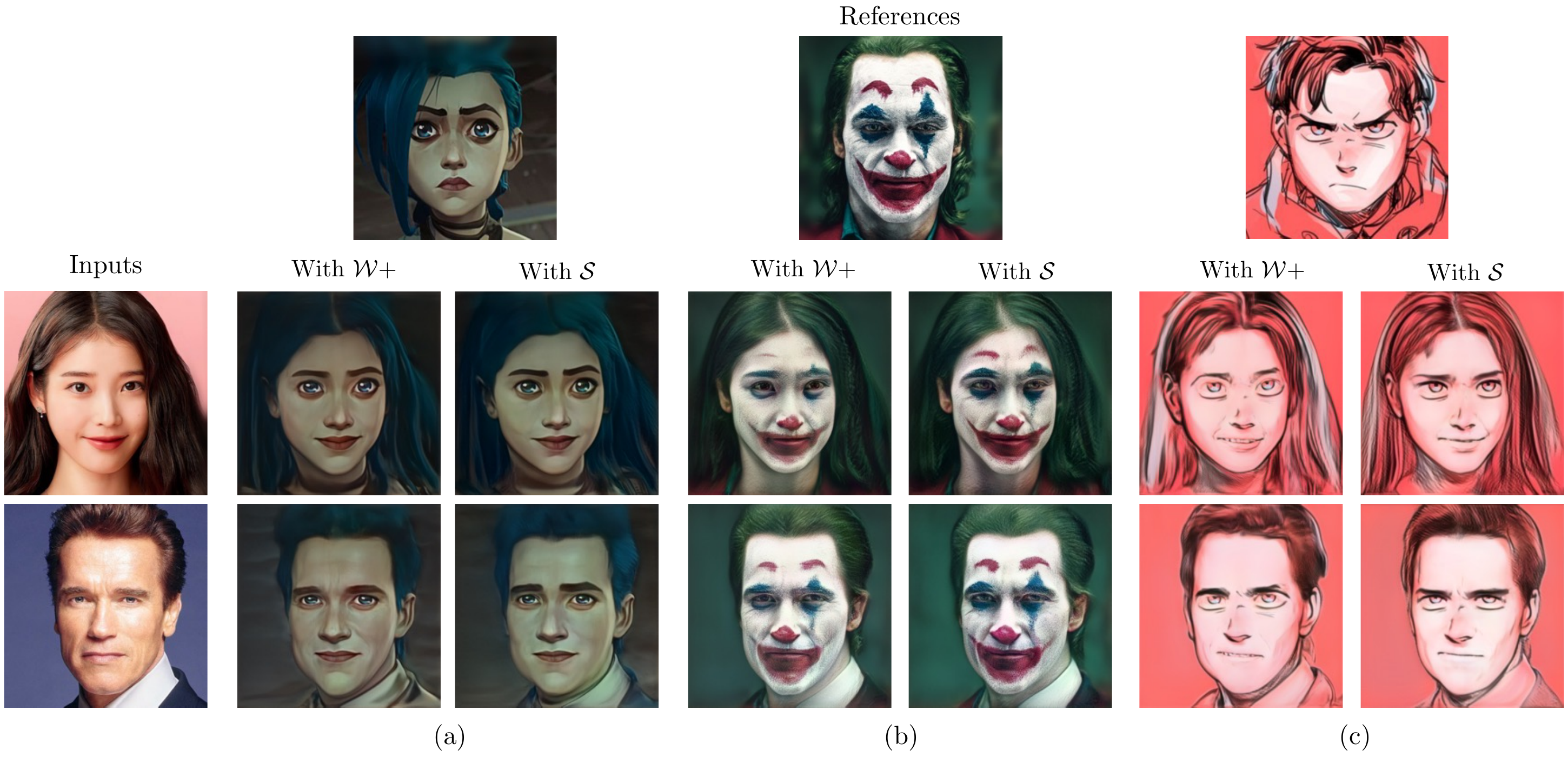}
    \caption{
    We study how the choice of latent space to do style mixing affects JoJoGAN.
    Style mixing in $\mathcal{S}$ space gives more accurate color reproduction in (a) and (b) and better stylization effect (note the eyes) in (c).
    }
    \label{fig:style_mix_p}
    \vspace{\figmargin}
\end{figure}
Style mixing in Equation~\eqref{eq:sample_w} allows us to generate more paired datapoints.
It is reasonable to map faces with slight difference in textures, colors, to the same reference.
As such it is pertinent that while we style mix to generate different faces, we need certain features such as identity, face pose, etc to remain the same.
We study how the choice of latent space to do style mixing affects the stylization.
In \figref{style_mix_p} we see that style mixing in $\mathcal{S}$ gives better color reproduction and overall stylization effect.
This is because $\mathcal{S}$ is more disentangled~\cite{wu2021stylespace} and allows us to more aggressively style mix without changing the features we want intact.

\subsection{Varying dataset}\label{section:vary_dataset}
\begin{figure}[ht]
    \centering
    \includegraphics[width=1\linewidth]{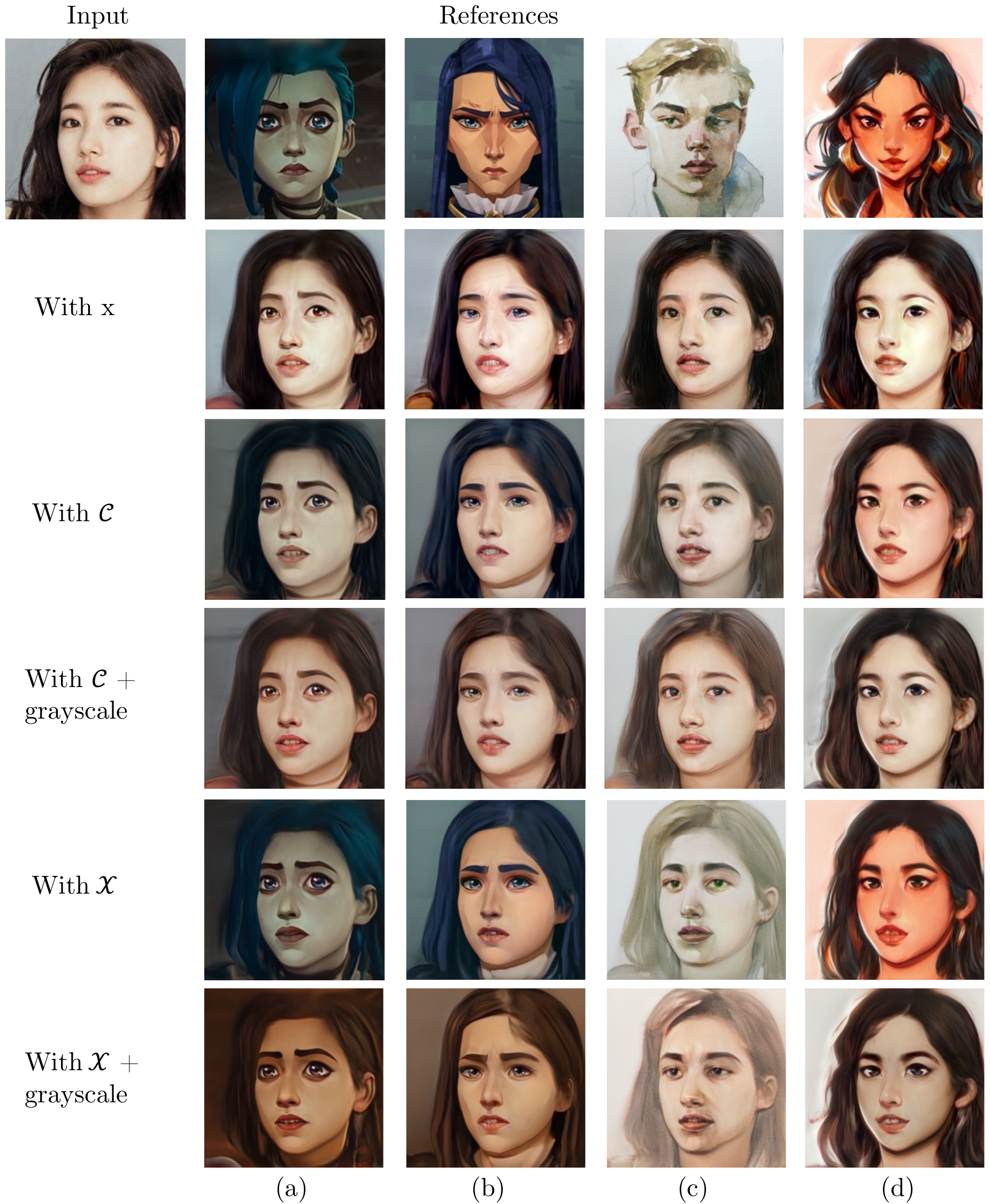}
    \caption{The choice of training data has an effect.
      {\bf First row:} when there is just one example in $\mathcal{W}$, JoJoGAN transfers relatively little style,
      likely because it is trained to map ``few'' images to the stylized example.
      {\bf Second row:} same training procedure as in \figref{preserve_color}, using $\mathcal{C}$.
      {\bf Third row:} same training procedure as second row but with grayscale images for Equation~\eqref{eq:loss}.
      {\bf Fourth row:} same training procedure as in \figref{preserve_color}, using $\mathcal{X}$.
    {\bf Fifth row:} same training procedure as Fourth row but with grayscale images for Equation~\eqref{eq:loss}.
    }
    \label{fig:preserve_color2}
    \vspace{\figmargin}
\end{figure}
Using $\mathcal{C}$ and $\mathcal{X}$ gives different stylization effects.
Finetuning with $\mathcal{X}$ accurately reproduces the color profile of the reference while $\mathcal{C}$ tries to preserve the input color profile.
However this is insufficient to fully preserve the colors as we see in \figref{preserve_color2}.
Grayscaling the images before computing the loss in Equation~\eqref{eq:loss} in addition to finetuning with $\mathcal{C}$ gives us stylization effects without altering the color profile.
We show that it is necessary to use both $\mathcal{C}$ and grayscaling to achieve this effect and using $\mathcal{X}$ and grayscaling is insufficient.

\subsection{Feature matching loss}\label{section:feat_loss}
For discriminator feature matching loss, we compute the intermediate activations after resblock $2, 4, 5, 6$.
\begin{figure*}[ht!]
    \centering
    \includegraphics[width=1\linewidth]{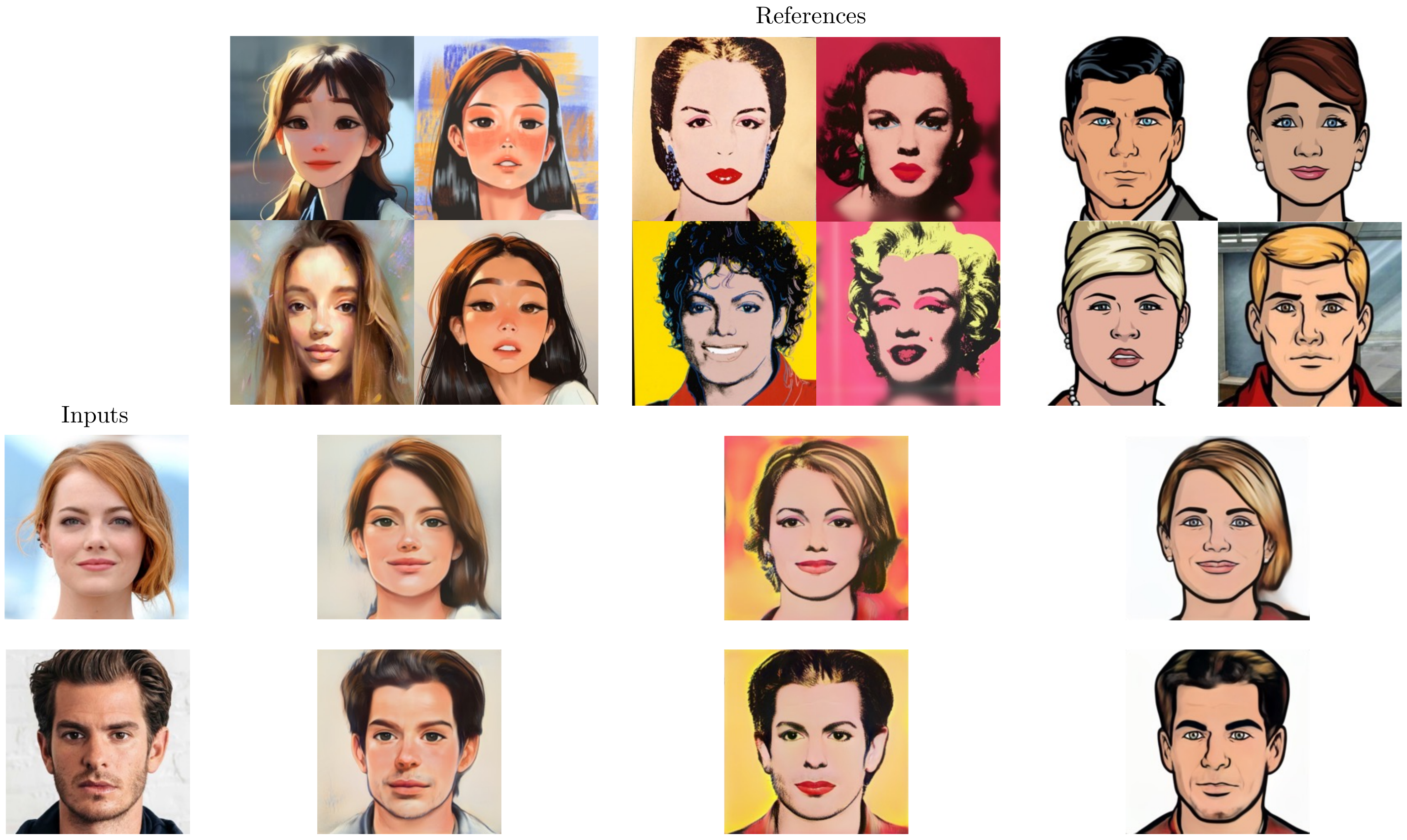}
    \caption{
    \textbf{More multi-shot examples}
    }
    \label{fig:multi2}
    \vspace{\figmargin}
\end{figure*}

\begin{figure}[ht]
    \centering
    \includegraphics[width=1\linewidth]{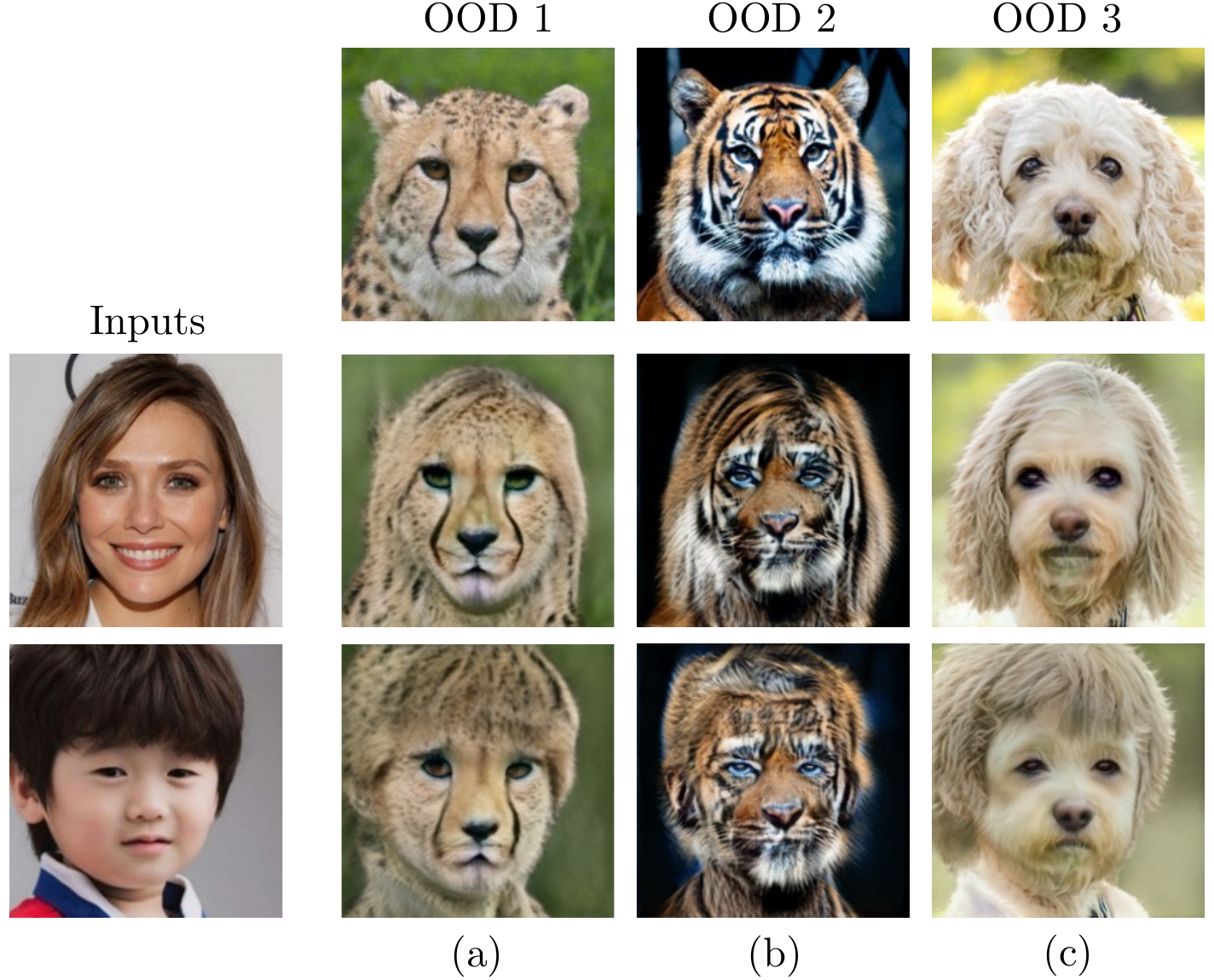}
    \caption{
    JoJoGAN produces unsatisfactory style transfers on OOD cases, producing human-animal hybrids.
    }
    \label{fig:oodw}
    \vspace{\figmargin}
\end{figure}
\begin{figure}[ht]
    \centering
    \includegraphics[height=1\textheight]{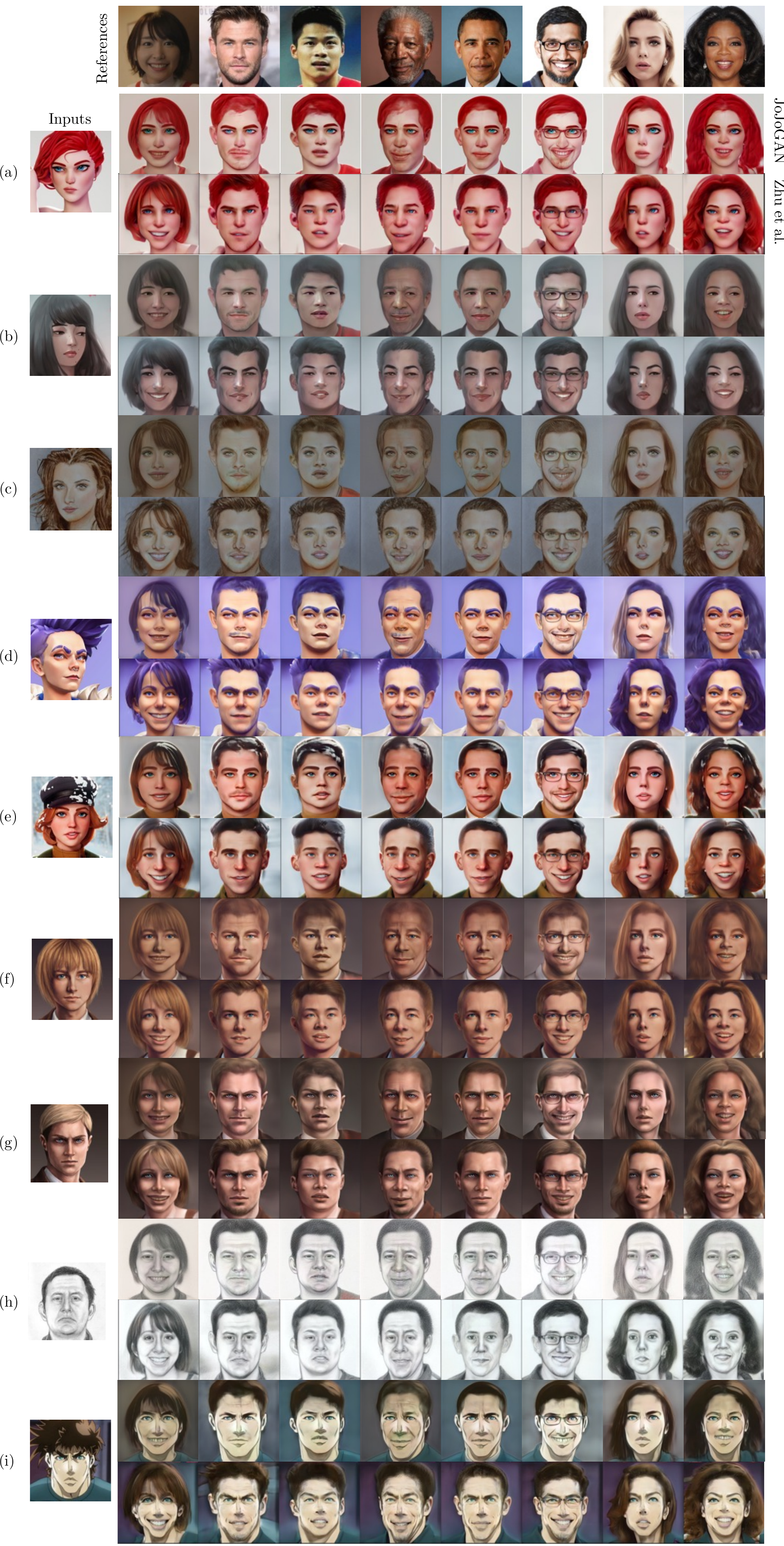}
    \caption{
      We compare with Zhu \etal~\cite{zhu2022mind} on all examples for references used in their paper and described as hard cases there. For each reference, the top row is JoJoGAN while the second row is Zhu \etal
      Note how their method distorts chin shape, while JoJoGAN produces strong outputs.
    }
    \label{fig:comparisons_zhu}
    \vspace{\figmargin}
\end{figure}
\begin{figure}[ht]
    \centering
    \includegraphics[width=1\linewidth]{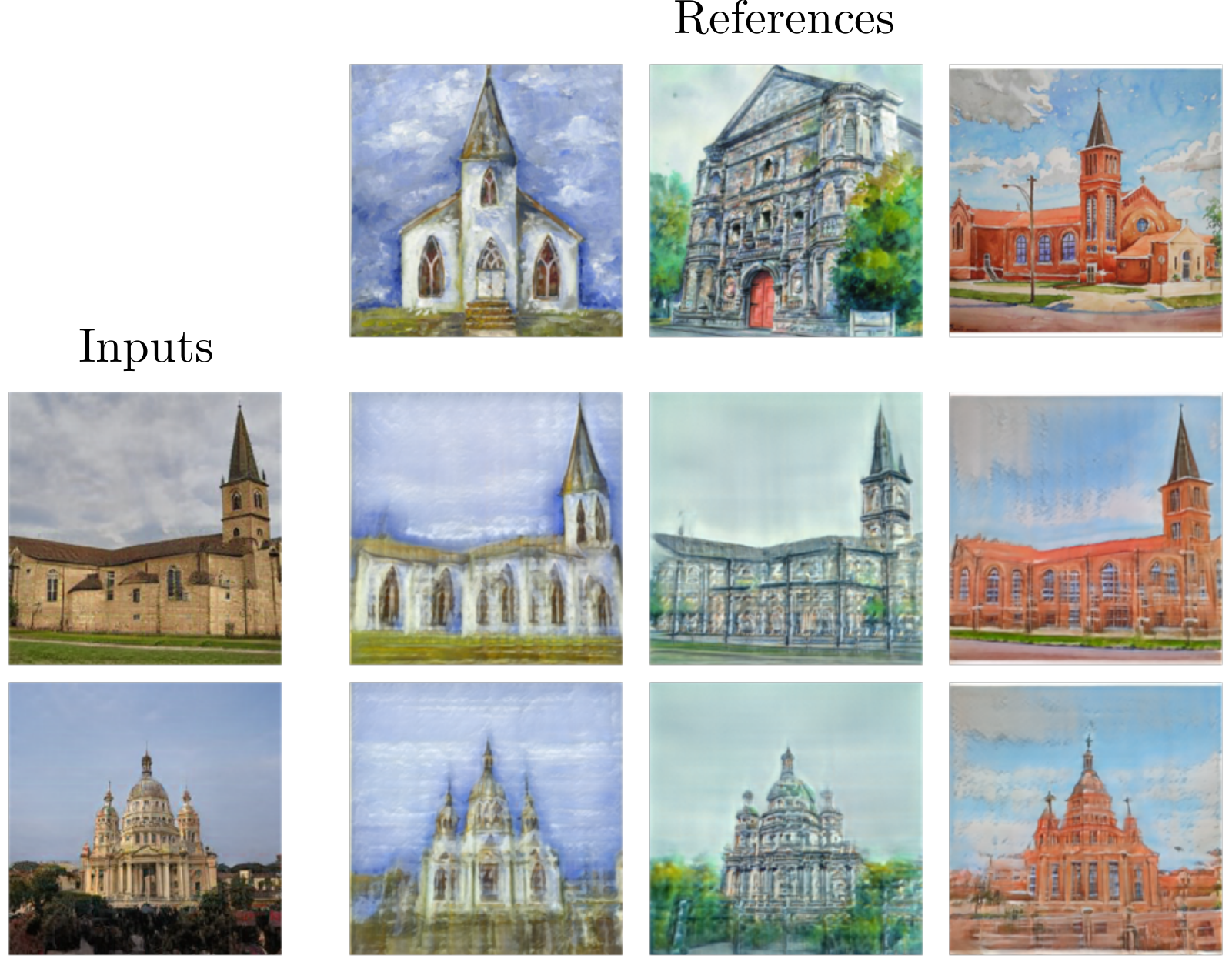}
    \caption{
      JoJoGAN is a method to benefit from what a StyleGAN knows, and so should apply to other
      domains where a well-trained StyleGAN is available.  Here we demonstrate JoJoGAN applied
      to LSUN-Churches.  
    }
    \label{fig:church}
    \vspace{\figmargin}
\end{figure}

\end{document}